%% file: pendu_arxiv.tex
\documentclass[letterpaper, 10 pt, conference]{ieeeconf}  
\usepackage{times}
\usepackage{courier}
\usepackage{epsfig}
\usepackage{graphicx}
\usepackage{amsmath}
\usepackage{amssymb, mathrsfs}
\usepackage{hyperref}
\usepackage{booktabs}
\usepackage[font=small,labelfont=bf]{caption}
\usepackage{url}
\usepackage[table,dvipsnames]{xcolor}
\usepackage{verbatim}
\usepackage{pygtex}
\usepackage{listing}
\usepackage{multirow}
\usepackage{algorithm}
\usepackage{algorithmic}

\definecolor{mygray}{gray}{0.8}
\DeclareMathOperator*{\E}{\mathbb{E}}
\DeclareMathOperator*{\maximize}{\scriptsize \mathbf{maximize}}

\IEEEoverridecommandlockouts                              

\title{\LARGE \bf
  Controlling an Inverted Pendulum with Policy Gradient Methods - A Tutorial 
}

\author{Swagat Kumar
  \thanks{* Swagat Kumar is  with the department of Computer Science, Edge Hill University,
  Ormskirk, UK L39 4QP. email: {\tt\small
  swagat.kumar@edgehill.ac.uk}} 
}

\begin{document}
	
	\maketitle
	\thispagestyle{empty}
	\pagestyle{empty}

	\begin{abstract}
     This paper provides details of implementing two important policy
     gradient methods to solve the OpenAI/Gym's pendulum problem.
     These are called Deep Deterministic Policy Gradient (DDPG) and
     the Proximal Policy Optimization (PPO) algorithms respectively.
     The problem is solved by using an actor-critic model where an
     actor network is used to learn the policy function and, a critic
     network to evaluates the action taken by the actor by learning to
     estimate the value or Q function. Apart from briefly explaining
     the mathematics behind these two algorithms, the details of
     implementing these algorithms using Python is provided which
     would help the readers in understanding the underlying complexity
     of these algorithms. In the process, the readers will be introduced
     to OpenAI/Gym, Tensorflow 2.x and Keras utilities used for
     implementing the above concepts. Through simulation experiments,
     it is shown that DDPG solves the problem in about 50-60 episodes
     and the PPO algorithm takes around 15-20 seasons for the same
     task. 
   \end{abstract} 
	
   \section{Introduction} \label{sec:intro} Reinforcement Learning
   (RL) \cite{sutton2018reinforcement} is a class of machine learning
   algorithms where an agent learns optimal behaviour through
   repetitive interaction with its environment that either rewards or
   penalizes for its actions. The agent learns the optimal behaviour
   by maximizing the cumulative future reward for a given task while
   finding a balance between \emph{exploration} (of new possibilities)
   and \emph{exploitation} (of past experience). This cumulative
   future reward (also known as the value or the Q-function) is
   generally not known apriori and is computed by using an approximate
   dynamic programming formulation starting with an initial estimate
   \cite{barto1995reinforcement}. Depending on how the value (or Q)
   function is estimated and policies are derived from them, the RL
   methods are broadly divided into two categories: (1) value-based
   methods and (2) policy-based methods.  The \emph{value-based}
   methods primarily aim at estimating the value or Q function which
   can then be used for deriving action policy by using a
   greedy approach. For discrete state and action spaces, the Q
   function is usually represented as a table of values, one for each
   state-action pair. This Q-table is then updated iteratively over
   time by using the Q-learning algorithm \cite{watkins1992q} while
   the agent is interacting with the environment. The Q-learning
   algorithm suffers from the \emph{curse-of-dimensionality} problem
   for discrete spaces leading to exponential computational time as
   the states and actions increase in dimension or range. The problems
   associated with discrete state spaces can be solved by using a deep
   neural network to approximate the Q-function which can now take
   continuous state values as inputs. The application of deep learning
   to solve RL problems has given rise to a field called \emph{deep
   reinforcement learning} \cite{li2017deep} that has been shown to
   solve many complex problems such as Atari games
   \cite{mnih2013playing}, robot control \cite{mnih2015human},
   autonomous driving \cite{wang2018deep} etc., with an intelligence
   that sometimes surpasses human experts \cite{holcomb2018overview}.
   The deep networks used for estimating Q function are also known as
   deep Q networks (DQN) that have been shown to provide amazing
   performance with concepts such as \emph{prioritized experience
   replay} (PER) \cite{schaul2015prioritized}, double DQN (DDQN)
   \cite{van2016deep}, dueling DQN \cite{sewak2019deep} etc. Most of
   these value-based methods still use discrete action spaces which
   limit their application to robot control problems that require
   continuous action values in terms of joint angle velocity or torque
   inputs. The problem of continuous action space is solved easily by
   using \emph{policy gradient} methods where the agent's policy is
   obtained by directly maximizing a given objective function through
   a gradient ascent algorithm \cite{peters2006policy}. The policy
   function itself can be approximated using a deep network whose
   parameters can be updated to maximize the output of another deep Q
   network (DQN) leading to a hybrid approach called the deep
   deterministic policy gradient (DDPG) that has been used for solving
   continuous robot control problem \cite{lillicrap2015continuous}.
   Since then, a number of major breakthroughs have been reported in
   the literature that has brought about a resurgence in this field
   attracting a large number of researchers in the recent past. One of
   the main advantages of reinforcement learning lies in the fact that
   the optimal behaviour of an autonomous agent can be learnt directly
   from its input-output data without having any apriori knowledge of
   the system models and their underlying dynamics. On the other hand,
   one of the main limitations of RL lies in the amount of data
   required for training the agent models.   

   The objective of this paper is to act as a tutorial for beginners
   making them easily grasp reinforcement learning concepts which are
   sometimes quite difficult to understand. Particularly, two policy
   gradient methods namely, deep deterministic policy gradient (DDPG)
   and proximal policy optimization (PPO) are described while solving
   the OpenAI/Gym's inverted pendulum problem. In the process, the
   readers are introduced to python programming with Tensorflow 2.x,
   Keras, OpenAI/Gym APIs. Readers interested in understanding and
   implementing DQN and its variants are advised to refer to
   \cite{kumar2020balancing} for a similar treatment on these topics.
   The rest of this paper is organized as follows. The problem
   definition and mathematical background for the above two policy
   gradient methods is discussed in Section \ref{sec:meth}. The
   experimental details and analysis of results is discussed in
   Section \ref{sec:expt} followed by conclusion in Section
   \ref{sec:conc}.

\begin{figure}[!t] \centering \begin{tabular}{ccc}
    \includegraphics[width=2cm, height=3cm]{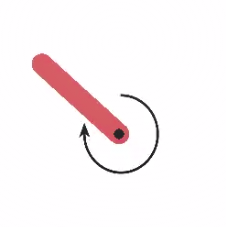}  &
    \includegraphics[width=2cm, height=3cm]{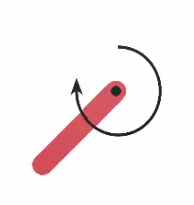} &
    \includegraphics[width=2cm, height=3cm]{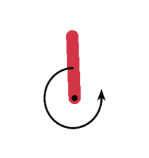} \\
    \small{(a)} & \small{(b)} & \small{(c)} \end{tabular}
  \caption{Various states of an inverted Pendulum. (a) and (b) are the
    intermediate unstable states; (c) is the final upright state which
  is desired.}
  \label{fig:pendu}
\end{figure}

   \section{Methods} \label{sec:meth}

   \subsection{The System}
   An inverted pendulum is a classical control problem having two
   degrees of freedom of motion and only one actuator to control its
   position. The motor is connected at the tip of the pendulum bar.
   The base joint of this bar is a free joint without any motor. The
   goal is to remain at zero angle (vertical position) with least
   rotational velocity and least effort. Some of the states of the
   pendulum is shown in Figure \ref{fig:pendu}. The vertical upright
   position (c) is the desirable state for the system. The range of
   various input and output parameters for this simulation environment
   is provided in Table \ref{tab:pendu_param}. A simple python code to
   visualize the dynamic simulation of the system is provided in code listing
   \ref{lst:pendu}. The render option should be disabled during actual
   training to speed up the computation.  Each episode involves 200
   iteration steps. The total reward for an episode is the sum of
   individual rewards obtained at each of these steps. Since, no
   training is involved in this code, the total reward for each of
   these episodes will be a large negative number. The problem is
   considered solved if the average reward for an episode remains
   above -200 for a considerable amount of time (say, 50 episodes). 

\begin{listing}
  \input{./pygtex_files/pendu_gym_test-py}
  \caption{\small Discretizing continuous states into discrete states}
  \label{lst:pendu}
\end{listing}

   \begin{table}[!htbp]
     \resizebox{\columnwidth}{!}{
     \begin{tabular}{|c|c|c|c|c|} \hline
       Variable Name & shape & Elements &  Min & Max \\ \hline
       \multirow{3}{*}{Observation (state), $s$} & \multirow{3}{*}{(3,1)} & $\cos{\theta}$ & -1.0 & 1.0 \\ \cline{3-5}
       &       & $\sin{\theta}$ & -1.0 & 1.0 \\ \cline{3-5}
       &       & $\dot{\theta}$ & -8.0 & 8.0 \\\hline
       Action, $a$         & (1,)  & Joint Effort   & -2.0 & 2.0 \\ \hline
       Reward, $r$     & (1,)  & $-\theta^2+0.1\dot{\theta}^2+0.001a^2$ &-16.273 & 0\\\hline
     \end{tabular}}
     \caption{Input-Output variables for `\texttt{Pendulum-v0}' Gym Simulation Environment.}
     \label{tab:pendu_param}
   \end{table}

   \subsection{Actor-Critic Model}\label{sec:ac} The Pendulum problem
   is solved in this paper using an \emph{Actor-Critic} model. In this
   model, an actor network is used to approximate the policy function
   $\mu_\theta(s)$ whereas a critic network is used to approximate the
   value function $V_\phi(s)$ or the function $Q_\phi(s,a)$. The
   actor-critic model allows one to implement separate training
   algorithm for the actor and the critic network and hence, provides
   greater flexibility compared to other models. The actor and critic
   class templates are shown in the code listings \ref{lst:actor} and
   \ref{lst:critic} respectively. The class template for the
   actor-critic agent is provided in listing \ref{lst:agent}.
   Following the double DQN formalism \cite{van2016deep}, a separate
   target network is created for both DDPG actor and critic models
   which share the same weights as that of the actual models in the
   beginning. The weights of the target networks are updated slowly
   during the training process by using Polyak Averaging
   \cite{polyak1992acceleration} as will be explained in the next
   section. The actor model for PPO algorithm returns mean and
   standard deviation of the policy distribution. The reinforcement
   learning agent initializes an actor model, a critic model and a
   buffer model. The buffer model is used for storing the experiences
   which will be used for training the DDPG agent. The agent provides
   two main functions - one for selecting policy (or action) and the
   other for training the agent. These functions are defined and
   explained in the later sections. In this paper, we only show the
   implementation of two main algorithms - DDPG and PPO. However, the
   same template could be used for incorporating other algorithms as
   well. 

   \begin{listing}
     \input{./pygtex_files/actor-py}
     \caption{Class template for Actor Network}
     \label{lst:actor}
   \end{listing}

   \begin{listing}
     \input{./pygtex_files/critic-py}
     \caption{Class template for the Critic Network}
     \label{lst:critic}
   \end{listing}

   \begin{listing}
     \input{./pygtex_files/agent-py}
     \caption{Class template for the actor-critic agent}
     \label{lst:agent}
   \end{listing}

   \subsection{Policy Gradient Methods}\label{sec:pgm}
   The policy gradient methods aims at learning a policy function
   $\pi_\theta=P(a|s)$ that maximizes an objective function $J(\theta)$ that
   computes the expected cumulative discounted rewards given by:
   \begin{equation}
     J(\theta) = \E\left[\sum_{t=0}^\infty r_{t+1}\right] 
     \label{eq:obj1}
   \end{equation}
   where $r_{t+1} = R(s_t, a_t)$ is the reward received by performing
   an action $a_t$ at state $s_t$. This can be expressed in terms of
   policy function $\pi_\theta(a|s)$ as follows:

   \begin{equation}
         J(\theta) = \E\left[\displaystyle
         \sum_{s\in\mathcal{S}}d^\pi(s)
       \sum_{a\in\mathcal{A}}\pi_\theta(a|s) Q^\pi(s,a) \right]
       \label{eq:obj2}
     \end{equation}
     where $d^\pi(s)$ is the stationary distribution of Markov chain for $\pi_\theta$ given by
     \begin{equation}
       d^\pi(s) = \lim_{t\rightarrow \infty} P(s_t=s|s_0,\pi_\theta)
       \label{eq:dpi}
     \end{equation}

  The policy parameter $\theta$ is updated using gradient ascent
  algorithm that aims to maximize the above objective function. The
  parameter update rule can be written as:
  \begin{equation}
    \theta = \theta + \eta \nabla_\theta J(\theta)
    \label{eq:sgd}
  \end{equation}
  
  The gradient of the objective function is given by:
  \begin{equation}
    \nabla_\theta J(\theta) = \E\left[\sum_{t=0}^\infty  Q^\pi(s_t,a_t)\nabla_\theta \log \pi_\theta(a_t|s_t)\right]
    \label{eq:j_grad}
  \end{equation}
  In terms of advantage function $A^\pi(s_t,
  a_t)=Q^\pi(s_t,a_t)-V^\pi(s_t)$, the above expression can be written
  as
  \begin{equation}
    \nabla_\theta J(\theta) = \E\left[\sum_{t=0}^\infty A^\pi(s_t, a_t)\nabla_\theta \log \pi_\theta(a_t|s_t)\right] 
    \label{eq:jgrad_adv}
  \end{equation}
  The advantage function can be estimated from the value function
  estimates (details are available in \cite{schulman2015high}) by
  using the generalized advantage estimator given by the following
  equation:
    \begin{equation}
    \hat{A}_t = \sum_{l=0}^\infty (\gamma \lambda)^l \delta_{t+l}
    \label{eq:gae}
  \end{equation}
  where parameter $0\le\lambda \le 1$ controls the trade-off between
  bias and variance and $\delta_t$ represents the time delay error
  given by
  \begin{equation}
    \delta_t = r_t + \gamma V(s_{t+1}) - V(s_t)
    \label{eq:tde}
  \end{equation}

   \begin{listing}
     \input{./pygtex_files/gae-py}
     \caption{Function definition for implementing generalized
     advantage estimates (GAE).}
     \label{lst:gae}
   \end{listing}

  As explained in the previous subsection, the above optimization
  problem is solved in this paper by using an \emph{actor-critic}
  method that combines the benefit of both value-based and
  policy-based methods. It consists of two models: Actor and Critic.
  The actor model is used to approximate the policy function
  $\pi_\theta(a|s)$ that computes an action for a given state while
  the critic model is used to evaluate this action by estimating the Q
  function $Q_\phi(s,a)$ or the value function $V_\phi(s)$.
  
  The critic model updates the q-value function parameter $\phi$ of
  the action value $Q_\phi(s,a)$ or the state-value $V_\phi(s)$
  function by using the time-delay error $\delta_t$ given by equation
  \eqref{eq:tde}. The weight update rule for the critic may be written
  as:

  \begin{equation} \phi \leftarrow \phi + \alpha_\phi \delta_t \nabla_\phi Q_\phi(s,a)
    \label{eq:critup} \end{equation}
  
  On the other hand, the actor model updates the policy parameter
  $\theta$ for $\pi_\theta(a|s)$ so as to maximize the value function
  estimated by the critic. This weight update rule for the actor may
  be written as 

  \begin{equation} \theta \leftarrow \theta +
    \alpha_\theta Q_\phi(s,a)\nabla_\theta \log \pi_\theta(a|s)
    \label{eq:actup} \end{equation}

   \subsection{Proximal Policy Optimization (PPO)}\label{sec:ppo}
   To improve the training stability, the parameter updates that
   result in large change in policy should be avoided. Trust Region
   Policy Optimization (TRPO) \cite{schulman2015trust} ensures this by
   enforcing KL divergence constraint on the size of the policy update
   at each iteration. This is achieved by optimizing the following objective function:

   \begin{equation}
     J(\theta)=\E_{\mathbf{s}\sim\rho^{\pi_{\theta_{\text{old}}}}, a\sim \pi_{\theta_{\text{old}}}} 
     \left[\frac{ \pi_\theta(a|s)}{\pi_{\theta_{\text{old}}}(a|s)} 
     \hat{A}_{\theta_{\text{old}}}(s,a) \right]
     \label{eq:j_trpo}
   \end{equation}
   subject to the trust region constraint given by:
   \begin{equation}
     \E_{\mathbf{s}\sim\rho^{\pi_{\theta_{\text{old}}}}}
     [D_{KL}(\pi_{\theta_{\text{old}}}(.|s)||\pi_\theta(.|s)] \le
     \delta
     \label{eq:kld} 
   \end{equation}
   where $\rho^{\pi_{\theta_{\text{old}}}}$ represents the state
   distribution under the old policy. The Proximal Policy Optimization
   (PPO) simplifies TRPO by using a clipped surrogate objective
   function while retaining similar performance. 
   
   Let us denote the
   probability ratio between new and old policy as:
   \begin{equation}
     r(\theta) = \frac{\pi_\theta(a|s)}{\pi_{\theta_{\text{old}}}(a|s)}
     \label{eq:prob_ratio}
   \end{equation}
   Then the objective function for TRPO becomes:
   \begin{equation}
     J^{TRPO}(\theta) = \E[r(\theta)\hat{A}_{\theta_{\text{old}}}(s,a)]
     \label{eq:trpo_obj}
   \end{equation}
   Maximizing this objective function without any restriction on
   distance between $\theta_{\text{old}}$ and $\theta$ would lead to
   instability with large parameter updates and big policy ratios. PPO
   avoids this by forcing $r(\theta)$ to stay within a small interval
   around 1, $[1-\epsilon, 1+\epsilon]$, where $\epsilon$ is a
   hyperparameter. The modified cost function maximized in PPO is
   given by
   \begin{align}
     J^{CLIP}(\theta) &= \E[\min(r(\theta) \hat{A}_{\theta_{\text{old}}}(s,a), \nonumber \\
     & \quad\quad  \text{clip}(r(\theta), 1-\epsilon, 1+\epsilon)\hat{A}_{\theta_{\text{old}}}(s,a))] \nonumber \\
     &= \mathscr{L}(s,a,\theta_{\text{old}}, \theta)
   \end{align}
   While applying PPO to an actor-critic kind of model, the above
   objective function is augmented with an error term on the value
   estimation and an entropy term as shown below for better
   performance:
   \begin{multline}
     J^{CLIP}(\theta) = \E[J^{CLIP}(\theta)  - c_1(V_\theta(s)  - V_{\text{target}})^2 \\ + c_2 H(s, \pi_\theta(.))]
     \label{eq:jclip2}
   \end{multline}

   The algorithm for the PPO-Clip method is provided in the algorithm
   listing \ref{alg:clip}. It is also possible to impose this
   constraint by including a penalty term on the KL divergence term
   $D_{KL}(\theta||\theta_{\text{old}})$ in the objective function
   which will ensure that the new policy does not deviate too much
   from the current policy. Hence the policy optimization problem given in equation \eqref{eq:j_trpo} and \eqref{eq:kld} may be rewritten as:
   \begin{multline}
     \maximize_\theta \left(\E[\frac{\pi_\theta(a|s)}{\pi_{\theta_{\text{old}}}(a|s)}\hat{A}] - 
     \beta \E[D_{KL}(\theta||\theta_{\text{old}})]\right)
     \label{eq:klp_obj}
   \end{multline}
   where $\beta$ controls the weight of the penalty which itself can
   be adaptively varied as the training progresses. The resulting PPO
   algorithm is shown in Algorithm listing \ref{alg:klp}.

   \begin{algorithm}[t]
     \caption{\small PPO-Clip Algorithm} 
      \label{alg:clip}
      \scriptsize
     \begin{algorithmic}[1]
       \REQUIRE initial policy parameters $\theta_0$, initial value function parameters $\phi_0$
       \FOR {$k=0,1,2, \dots$}
       \STATE Collect a set of trajectories $\mathscr{D}_k =
       \{\tau_i\}$ by running policy $\pi_k=\pi(\theta_k)$ in the
       environment.
       \STATE Compute rewards-to-go $\hat{R}_k$
       \STATE Compute advantage estimates $\hat{A}_k$ based on the current value function $V(\phi_k)$
       \STATE Update the policy by maximizing the PPO-Clip objective via stochastic gradient ascent:
       \begin{multline}
       \theta_{k+1} = \arg\max_\theta \frac{1}{|\mathscr{D}_k|T} \sum_{\tau\in\mathscr{D}_k} \sum_{t=0}^T  \\
       \min \left(r(\theta)A^{\pi_k}(s_t, a_t), g(\epsilon, A^{\pi_k}(s_t,a_t))\right)
       \label{eq:act_up}
       \end{multline}
       where 
       \[g(\epsilon,a) = \left\{\begin{array}{l}
                           (1+\epsilon)A\ \text{if}\ A\ge 0 \\
                           (1-\epsilon)A\ \text{if}\ A<0 
                       \end{array} \right.\]
       \STATE Update value function to minimize the following error function by using gradient descent algorithm:
       \begin{equation}
         \phi_{k+1} = \arg\min_\phi \frac{1}{|\mathscr{D}_k|T}\sum_{\tau \in\mathscr{D}_k}\sum_{t=0}^T
         \left(V_\phi(s_t) - \hat{R}_t\right)^2
         \label{eq:crit_up}
       \end{equation}
       \ENDFOR
     \end{algorithmic}
   \end{algorithm}

   \begin{algorithm}[H]
     \caption{\small PPO with Adaptive KL Penalty}
     \label{alg:klp}
     \scriptsize
     \begin{algorithmic}[1]
       \REQUIRE initial policy parameter $\theta_0$, initial value parameter $\phi_0$, initial KL penalty $\beta_0$, target KL-divergence $\delta$
       \FOR {$k=0,1,2,\dots$}
       \STATE Collect a set of partial trajectories $\mathscr{D}_k$ on policy $\pi=\pi(\theta_k)$.
       \STATE Estimate Advantage $\hat{A}_k$ based on current value function $V(\phi_k)$
       \STATE Compute policy update
       \[\theta_{k+1} = \arg\max_\theta \mathscr{L}_{\theta_k}(\theta) - \beta_k\bar{D}_{KL}(\theta||\theta_k)\]
       by taking k steps of mini-batch SGD
       \IF {$\bar{D}_{KL}(\theta_{k+1}||\theta_k)\ge 1.5\delta$}
       \STATE $\beta_{k+1} = 2 \beta_k$
       \ELSIF {$\bar{D}_{KL}(\theta_{k+1}||\theta_k)\le 1.5/\delta$}
       \STATE $\beta_{k+1} = \beta_k/2$ 
       \ENDIF
       \ENDFOR
     \end{algorithmic}
   \end{algorithm}

   The corresponding function definitions required for implementing
   PPO algorithm is provided in listings \ref{lst:ppo_train} and
   \ref{lst:ppo_agent_train} respectively. The adaptive KL penalty
   method is enabled by using the value `\texttt{penalty}' and the
   clip method is enabled by using the value `\texttt{clip}' for the
   flag `\texttt{method}' in the actor class object. The experiences
   are first collected in a buffer and then used
   for computing advantages by using the Generalized Advantage
   Estimator (GAE) equation \eqref{eq:gae}. These experiences and
   estimates are then split into a number of mini-batches which are
   then used for training actor and critic networks for a certain
   number of epochs. The target values are the cumulative discounted
   reward computed using a low pass filter function available with
   `\texttt{scipy.signal}' module. The
   `\texttt{tensorflow\_probability}' module is used for computing
   likelihood ratio, KL divergence and the action probability
   distribution. The buffer is cleared or emptied after each training
   step. This makes the PPO algorithm an \emph{on-policy} method that uses
   the current experiences to train the model. 

   \begin{listing}
     \input{./pygtex_files/ppo_train-py}
     \caption{Training Actor and Critic using PPO Algorithm}
     \label{lst:ppo_train}
   \end{listing}

   \begin{listing}
     \input{./pygtex_files/ppo_agent_train-py}
     \caption{PPO training algorithm for the Agent}
     \label{lst:ppo_agent_train}
   \end{listing}

  \subsection{Deep Deterministic Policy Gradient (DDPG) Algorithm} \label{sec:ddpg}

  Deep Deterministic Policy Gradient (DDPG)  algorithm
  \cite{lillicrap2015continuous} concurrently learns a Q-function and
  a policy. It uses off-policy data and the Bellman equation to learn
  the Q-function and then learns a policy by maximizing the Q-function
  through a gradient ascent algorithm. This allows DDPG to apply deep
  Q-learning to continuous action spaces. 

The Q-learning in DDPG is performed by minimizing the following mean
squared bellman error (MBSE) loss with stochastic gradient descent:
\begin{multline} L(\phi, \mathscr{D}) = \E_{(s,a,r,s',d)\sim
    \mathscr{D}}\biggl[ \biggl( Q_\phi(s,a) -  \\
      (r + \gamma(1-d)Q_{\phi_{\text{targ}}}(s',
  \mu_{\theta_{\text{targ}}}(s')))\biggr)^2\biggr] \\
  =  \E_{(s,a,r,s',d)\sim \mathscr{D}} [ Q_\phi(s,a) - y(r,s',d)]
  \label{eq:ddpg_qloss}
\end{multline}
where $\mu_{\theta_{\text{targ}}}$ is the target policy,
$Q_{\phi_{\text{targ}}}$ is the target Q network and $y(r,s',d)$ is
the target critic value required for training the Q network. 

The policy learning in DDPG aims at learning a deterministic policy
$\mu_\theta(s)$ which gives action that maximizes $Q_\phi(s,a)$. Since
the action space is continuous, it is assumed that the Q function is
differentiable with respect to action. The policy parameters $\theta$
are updated by performing gradient ascent to solve the following
optimization problem:
\begin{equation}
  \max_\theta \E_{s\sim\mathscr{D}} [Q_\phi(s, \mu_\theta(s))]
  \label{eq:ddpg_po}
\end{equation}
 The parameters of the target networks are updated slowly compared to
 the main network by using Polyak Averaging \cite{polyak1992acceleration} as shown below:
\begin{eqnarray}
  \phi_{\text{targ}} &\leftarrow& \rho\phi_{\text{targ}} + (1-\rho)\phi \nonumber \\
  \theta_{\text{targ}} &\leftarrow& \theta \phi_{\text{targ}} + (1-\rho)\theta \nonumber
  \label{eq:pa}
\end{eqnarray}
where $0\le \rho \le 1$ is a hyper parameter. 

The pseudocode for DDPG algorithm is provided in the Algorithm table
\ref{alg:ddpg} and its python implementation is provided in listing
\ref{lst:ddpg}. The DDPG agent samples a batch of experiences from the
replay buffer and uses it to train the actor and critic networks. As
explained above, the actor implements gradient ascent algorithm by
using a negative of the gradient term computed using Tensorflow's
`\texttt{GradientTape}' utility. The critic, on the other hand, learns
by applying gradient descent to minimize the time-delay Q-function
error. 

\begin{algorithm}[H]
  \small
  \caption{\small DDPG Algorithm}
  \label{alg:ddpg}
  \scriptsize
  \begin{algorithmic}[1]
    \STATE input: initial policy parameter $\theta$, initial Q-function parameters $\phi$, empty replay buffer $\mathscr{D}$.
    \STATE set target parameters equal to the main parameters:
    $\theta_{\text{targ}}\leftarrow \theta$,
    $\phi_{\text{targ}}\leftarrow \phi$. 
    \REPEAT
    \STATE Observe state $s$ and select action
    $a=\text{clip}(\mu_\theta(s)+\epsilon, a_{Low}, a_{High})$, where
    $\epsilon \sim \mathscr{N}$. 
    \STATE Execute the action $a$ in the environment and obtain next
    state $s'$, reward $r$, done signal $d$ to indicate if it is a
    terminal state. 
    \STATE Store $(s,a,r,s',d)$ in the replay buffer $\mathscr{D}$
    \STATE if $s'$ is terminal, reset the environment.
    \IF {it's time to update}
    \FOR {for a certain number of steps}
    \STATE Randomly sample a batch of transitions $B={(s,a,r,s',d)}$ from $\mathscr{D}$.
    \STATE Compute targets: $y(r,s',d) = r + \gamma (1-d)Q_{\phi_\text{targ}}(s', \mu_{\theta_\text{target}}(s'))$
    \STATE Update Q function by applying one step of gradient descent using
    $\displaystyle \nabla_\phi \frac{1}{|B|}\sum_{(s,a,r,s',d) \in B}(Q_\phi(s,a) - y(r,s',d))^2$ 
    \STATE Update policy by one step of gradient ascent using
    $\displaystyle \nabla_\theta \frac{1}{|B|}\sum_{s\in B} Q_\phi(s,\mu_\theta(s))$
    \STATE Update target networks using Polyak Averaging given by equations \eqref{eq:pa}
    \ENDFOR
    \ENDIF
    \UNTIL convergence
  \end{algorithmic}
\end{algorithm}

   \begin{listing}
     \input{./pygtex_files/ddpg_train-py}
     \caption{\small Function Definitions for implementing DDPG algorithm}
     \label{lst:ddpg}
   \end{listing}


   \section{Other Implementation Details}

   The main function for implementing these two algorithms is shown in
   the listing \ref{lst:main}. The main training loop involves the
   following steps: (1) Compute action as per the current agent policy
   (2) Obtain reward from the environment by executing the above
   action, (3) Store this experience tuple $(s,a,r,s',d)$ in a buffer.
   (4) Use this experience to the train the agent. The environment is
   reset once the terminal state is reached indicating the end of
   the current episode. The PPO algorithm is  an \emph{on-policy}
   method and hence, the experiences in the buffer are discarded after
   each training iteration. On the other hand, DDPG is an
   \emph{off-policy} method that uses a sample of past experiences to
   train the agent. 
   \begin{listing}
     \input{./pygtex_files/main-py}
     \caption{\small Main Python File}
     \label{lst:main}
   \end{listing}

   \begin{listing}
     \input{./pygtex_files/policy-py}
     \caption{\small Agent policy function for generating action for a given state}
     \label{lst:policy}
   \end{listing}

 The agent policy used for generating action for a given state is
 shown in listing \ref{lst:policy}. The actor in DDPG uses a noisy
 version of a deterministic policy to generate actions for a given
 state. On the other hand, the actor in PPO draws an action from a
 policy distribution.

   \begin{listing}
     \input{./pygtex_files/build_net-py}
     \caption{\small Deep Neural Network for Actor and Critic Models}
     \label{lst:build_net}
   \end{listing}
   
   Each of the actor and critic models use a deep network to estimate
   the policy and value functions respectively. Tensorflow's Keras
   APIs are used to create these deep networks as shown in the code
   listing \ref{lst:build_net}. The actor uses a sequential
   128-64-64-1 multiple layer perceptron (MLP) network to estimate the
   policy function $\mu_\theta$. Similarly, the critic model uses a
   sequential 64-64-64-1 MLP network to estimate the value function
   $V_\phi(s)$ for given state input. It is to be noted that the
   action and states are continuous floating point values for the
   current problem.

   \section{Experiments and Results}\label{sec:expt}
   The full source codes for the implementations are available on the
   GitHub \cite{skgit2021} for the convenience of readers. These codes
   are implemented in Python 3.x by using Tensorflow 2.x and Keras
   APIs. TF 2.0 uses eager implementation by default and hence
   provides cleaner interface compared to TF 1.0 that uses
   placeholders and sessions instead. These codes could be easily
   implemented on various online GPU cloud platforms such as Google
   Colab \cite{colab} and Kaggle \cite{kaggle}. 
   
\begin{figure}[htpb]
     \centering
     \includegraphics[scale=0.5]{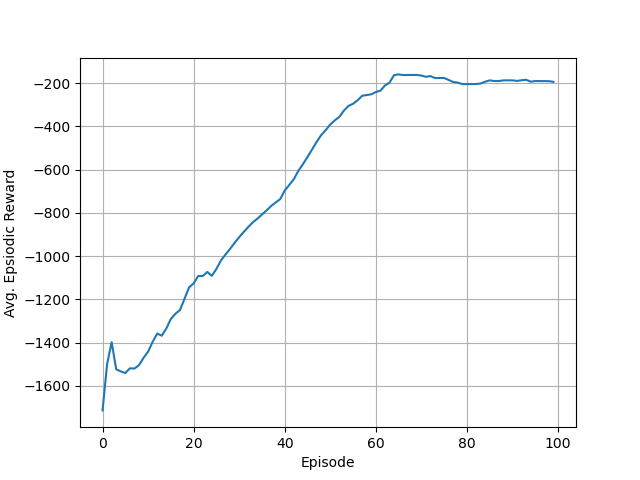}
     \caption{Training performance with DDPG algorithm. It solves the
     problem within 50-60 episodes.}     
     \label{fig:ddpg}
   \end{figure}

   The training results for DDPG algorithm is shown in Figure
   \ref{fig:ddpg}. It shows the average reward of last 40 episodes
   against the training episodes. The problem is considered solved
   when this average reward exceeds -200. As one can see in this
   figure, the DDPG algorithm solves the problem in about 50-60
   episodes. The complete execution takes about a couple of hours on
   Google Colab.  

\begin{figure}[htbp]
     \centering
     \includegraphics[scale=0.5]{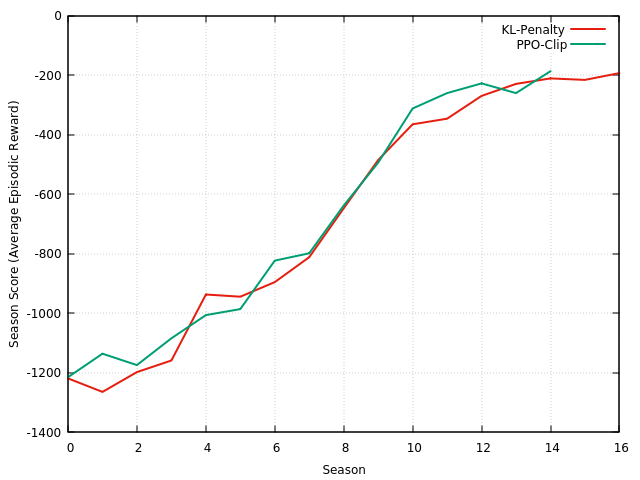}
     \caption{Training Performance of PPO Algorithm. Season score is
     the average episodic reward for a season. Each season consists of
   50 episodes.}
     \label{fig:ppo1}
   \end{figure}

\begin{figure}[htbp]
     \centering
     \begin{tabular}{cc}
       \hspace{-0.5cm}\includegraphics[scale=0.25]{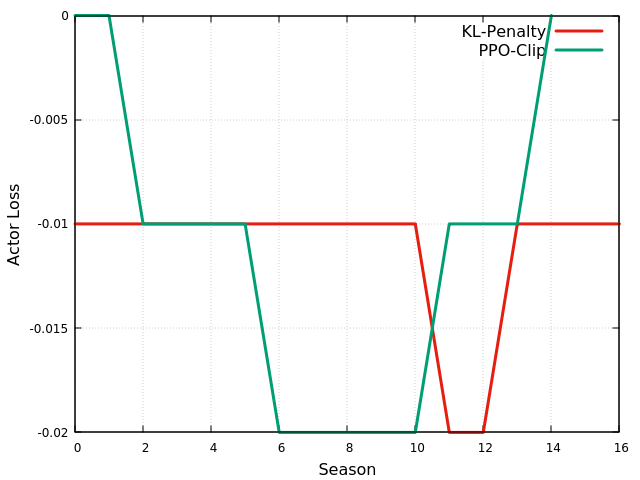} & 
     \includegraphics[scale=0.25]{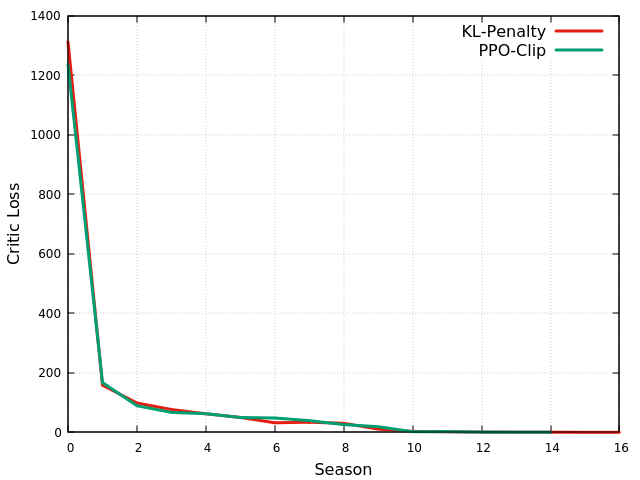}\\
     {\scriptsize (a)} & {\scriptsize (b)}\\
     \hspace{-0.5cm}\includegraphics[scale=0.25]{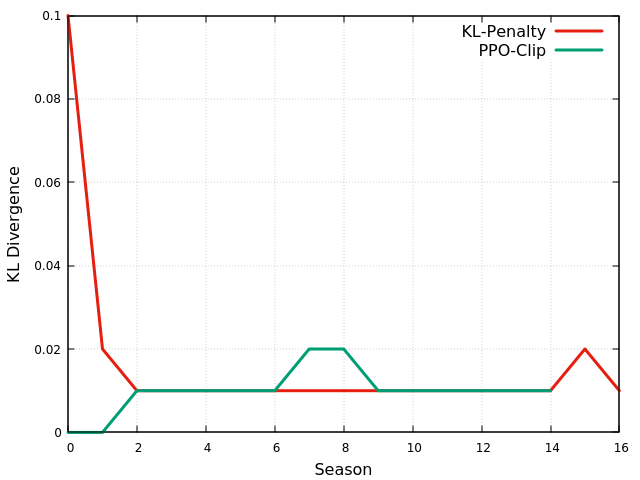} & 
     \includegraphics[scale=0.25]{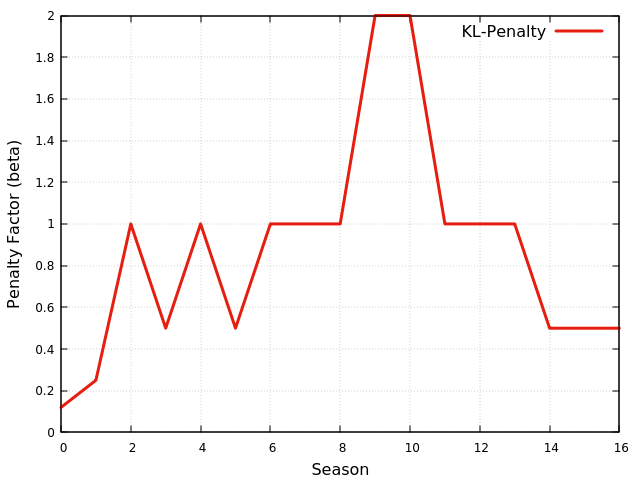}\\
     {\scriptsize (c)} & {\scriptsize (d) }
   \end{tabular}
     \caption{Training Performance of PPO algorithms: (a) Actor loss
     (b) Critic Loss (c) KL Divergence and (d) Penalty factor
   ($\beta$)}
     \label{fig:ppo2}
   \end{figure}

   The results for PPO algorithm is shown in Figure
   \ref{fig:ppo1}-\ref{fig:ppo2}. The Figure \ref{fig:ppo1} shows the
   average episodic reward of a given season as the training proceeds.
   Each season consists of about 50 episodes. The problem is
   considered solved if the season score exceeds a reward of -200. The
   actor and critic losses are shown in Figure \ref{fig:ppo2}(a)-(b).
   The critic loss decreases monotonically indicating that the network
   is able to minimize the time-delay error of the Q-network. The
   KL-Divergence between the old policy and the new policy is shown in
   Figure \ref{fig:ppo2}(c). It shows that the KL-divergence value
   remains close to the KL target value of 0.01. The variation of the
   KL penalty factor $\beta$ value in adaptive KL penalty method
   (Algorithm \ref{alg:klp}) is shown in Figure \ref{fig:ppo2}(d).  
   
   The hyper parameters used for these two algorithms is shown in
   Table \ref{tab:hp}. The number of samples used during each training
   iteration is about 10,000 as each episode has about 200 time-steps.
   The content of the buffer is discarded after each training
   iteration. On the other hand, the replay buffer of DDPG algorithm
   has a maximum size of 20,000 and the old experiences are removed as
   the buffer gets filled over time. Each training step in PPO
   involves multiple training epochs each of which uses a mini-batch
   of the experiences stored in the buffer. 

   \begin{table}[htbp]
     \centering
     \begin{tabular}{|p{3cm}|c|p{2cm}|c|}\hline  \hline
       \multicolumn{2}{|c|}{PPO} & \multicolumn{2}{c|}{DDPG} \\ \hline \hline
       Actor learning rate, $\alpha_a$ & 1e-4 & $\alpha_a$ & 1e-3 \\ \hline
       Critic learning rate, $\alpha_c$ & 2e-4 & $\alpha_c$ & 2e-3 \\ \hline
       Discount factor, $\gamma$ & 0.9 & $\gamma$ & 0.99 \\ \hline
       Discount factor in GAE, $\lambda$ & 0.95 & Polyak averaging factor, $\tau$ & 0.005 \\ \hline
       KL penalty factor, $\beta$ & 0.5 &  &  \\ \hline
       Clip factor, $\epsilon$ & 0.2 & & \\ \hline
       KL Target & 0.01 & & \\ \hline
       Training epochs & 20 & & \\ \hline
       Batch size & 200 & Batch size & 64   \\ \hline
       Buffer Size, $t_{\text{max}}$ & 10,000 &  Replay buffer size & 20,000 \\ \hline
     \end{tabular}
     \caption{Hyper-parameters used for implementing PPO and DDPG Algorithms}
     \label{tab:hp}
   \end{table}

   \subsection*{Discussion}
   It may seem that DDPG is faster than PPO in this case. However,
   making such direct comparison may not be fair. It is to be noted
   that DDPG is an off-policy method while PPO is an on-policy method.
   Secondly, PPO provides more efficient training compared to DDPG for
   more complex problems such Kuka-Grasp \cite{kukagrasp} or
   roboschool \cite{robos}.  Third, there are more efficient versions of
   actor-critic model such as A3C \cite{babaeizadeh2016ga3c} where
   data collection and training can be carried out by multiple workers
   working in parallel. Finally, the intention of this paper is not to
   make any comparison between these two algorithms. The objective is
   to understand the underlying concepts behind these algorithms and
   see them working on a simpler problem. 

   \section{Conclusion}\label{sec:conc}
   This paper provides the details of implementing two main policy
   gradient methods, namely, DDPG and PPO to solve the OpenAI/Gym's
   inverted pendulum problem. The mathematical background of these two
   methods are discussed along with their implementation using Python
   programming. The problem is solved by using an actor-critic model
   where the actor is used to estimated the agent's policy which is
   evaluated using the critic network. The information provided in
   this paper will be useful for beginners who would like to acquire
   advanced understanding of reinforcement learning concepts in
   minimum time. The actual code is made available on GitHub for the
   convenience of readers and can be easily implemented using online
   GPU platforms such as Google Colab \cite{colab} and Kaggle
   \cite{kaggle}. Results show that the DDPG algorithm takes around
   50-60 episodes to solve the problem while the PPO takes around
   15-20 seasons (about 1000 episodes) for the same.


\bibliographystyle{ieee} 
\bibliography{ref} 

\end{document}

%% file: pygtex_files/pendu_gym_test-py.tex
\begin{Verbatim}[commandchars=\\\{\},baselinestretch=0.9,frame=leftline,framesep=1.5ex,framerule=0.8pt,fontsize=\tiny]
\PY{k+kn}{import} \PY{n+nn}{gym}
\PY{n}{env} \PY{o}{=} \PY{n}{gym}\PY{o}{.}\PY{n}{make}\PY{p}{(}\PY{l+s+s2}{\PYZdq{}}\PY{l+s+s2}{Pendulum\PYZhy{}v0}\PY{l+s+s2}{\PYZdq{}}\PY{p}{)}
\PY{n}{state\PYZus{}size} \PY{o}{=} \PY{n}{env}\PY{o}{.}\PY{n}{observation\PYZus{}space}\PY{o}{.}\PY{n}{shape}
\PY{n}{action\PYZus{}size} \PY{o}{=} \PY{n}{env}\PY{o}{.}\PY{n}{action\PYZus{}space}\PY{o}{.}\PY{n}{shape}
\PY{n}{upper\PYZus{}bound} \PY{o}{=} \PY{n}{env}\PY{o}{.}\PY{n}{action\PYZus{}space}\PY{o}{.}\PY{n}{high}\PY{p}{[}\PY{l+m+mi}{0}\PY{p}{]}
\PY{n}{lower\PYZus{}bound} \PY{o}{=} \PY{n}{env}\PY{o}{.}\PY{n}{action\PYZus{}space}\PY{o}{.}\PY{n}{low}\PY{p}{[}\PY{l+m+mi}{0}\PY{p}{]}

\PY{k}{for} \PY{n}{ep} \PY{o+ow}{in} \PY{n+nb}{range}\PY{p}{(}\PY{l+m+mi}{20}\PY{p}{)}\PY{p}{:}
    \PY{n}{state} \PY{o}{=} \PY{n}{env}\PY{o}{.}\PY{n}{reset}\PY{p}{(}\PY{p}{)}
    \PY{n}{done} \PY{o}{=} \PY{n+nb+bp}{False}
    \PY{n}{ep\PYZus{}reward} \PY{o}{=} \PY{l+m+mi}{0}
    \PY{k}{while} \PY{o+ow}{not} \PY{n}{done}\PY{p}{:}
        \PY{n}{env}\PY{o}{.}\PY{n}{render}\PY{p}{(}\PY{p}{)}
        \PY{n}{action} \PY{o}{=} \PY{n}{env}\PY{o}{.}\PY{n}{action\PYZus{}space}\PY{o}{.}\PY{n}{sample}\PY{p}{(}\PY{p}{)}
        \PY{n}{next\PYZus{}state}\PY{p}{,} \PY{n}{reward}\PY{p}{,} \PY{n}{done}\PY{p}{,} \PY{n}{\PYZus{}} \PY{o}{=} \PY{n}{env}\PY{o}{.}\PY{n}{step}\PY{p}{(}\PY{n}{action}\PY{p}{)}
        \PY{n}{ep\PYZus{}reward} \PY{o}{+}\PY{o}{=} \PY{n}{reward}
        \PY{k}{if} \PY{n}{done}\PY{p}{:}
            \PY{k}{print}\PY{p}{(}\PY{l+s+s1}{\PYZsq{}}\PY{l+s+s1}{Episode: \PYZob{}\PYZcb{}, Reward: \PYZob{}\PYZcb{}}\PY{l+s+s1}{\PYZsq{}}\PY{o}{.}\PY{n}{format}\PY{p}{(}\PY{n}{ep}\PY{p}{,} \PY{n}{ep\PYZus{}reward}\PY{p}{)}\PY{p}{)}
            \PY{k}{break}
\PY{n}{env}\PY{o}{.}\PY{n}{close}\PY{p}{(}\PY{p}{)}
\end{Verbatim}

%% file: pygtex_files/actor-py.tex
\begin{Verbatim}[commandchars=\\\{\},baselinestretch=0.9,frame=leftline,framesep=1.5ex,framerule=0.8pt,fontsize=\tiny]
\PY{k}{class} \PY{n+nc}{Actor}\PY{p}{:}
    \PY{k}{def} \PY{n+nf+fm}{\PYZus{}\PYZus{}init\PYZus{}\PYZus{}}\PY{p}{(}\PY{n+nb+bp}{self}\PY{p}{,} \PY{n}{state\PYZus{}size}\PY{p}{,} \PY{n}{action\PYZus{}size}\PY{p}{)}\PY{p}{:}
        \PY{n+nb+bp}{self}\PY{o}{.}\PY{n}{model} \PY{o}{=} \PY{n+nb+bp}{self}\PY{o}{.}\PY{n}{\PYZus{}build\PYZus{}net}\PY{p}{(}\PY{p}{)}
        \PY{n+nb+bp}{self}\PY{o}{.}\PY{n}{optim} \PY{o}{=} \PY{n}{tf}\PY{o}{.}\PY{n}{keras}\PY{o}{.}\PY{n}{optimizers}\PY{o}{.}\PY{n}{Adam}\PY{p}{(}\PY{n+nb+bp}{self}\PY{o}{.}\PY{n}{lr}\PY{p}{)}

        \PY{k}{if} \PY{n+nb+bp}{self}\PY{o}{.}\PY{n}{flag} \PY{o}{==} \PY{l+s+s1}{\PYZsq{}}\PY{l+s+s1}{ddpg}\PY{l+s+s1}{\PYZsq{}}\PY{p}{:}
            \PY{n+nb+bp}{self}\PY{o}{.}\PY{n}{target} \PY{o}{=} \PY{n+nb+bp}{self}\PY{o}{.}\PY{n}{\PYZus{}build\PYZus{}net}\PY{p}{(}\PY{p}{)}
            \PY{n+nb+bp}{self}\PY{o}{.}\PY{n}{target}\PY{o}{.}\PY{n}{set\PYZus{}weights}\PY{p}{(}\PY{n+nb+bp}{self}\PY{o}{.}\PY{n}{model}\PY{o}{.}\PY{n}{get\PYZus{}weights}\PY{p}{(}\PY{p}{)}\PY{p}{)}
        \PY{k}{elif} \PY{n+nb+bp}{self}\PY{o}{.}\PY{n}{flag} \PY{o}{==} \PY{l+s+s1}{\PYZsq{}}\PY{l+s+s1}{ppo}\PY{l+s+s1}{\PYZsq{}}\PY{p}{:}
            \PY{n}{logstd} \PY{o}{=} \PY{n}{tf}\PY{o}{.}\PY{n}{Variable}\PY{p}{(}\PY{n}{np}\PY{o}{.}\PY{n}{zeros}\PY{p}{(}\PY{n}{shape}\PY{o}{=}\PY{n}{action\PYZus{}size}\PY{p}{,} \PY{n}{dtype}\PY{o}{=}\PY{n}{np}\PY{o}{.}\PY{n}{float32}\PY{p}{)}\PY{p}{)}
            \PY{n+nb+bp}{self}\PY{o}{.}\PY{n}{model}\PY{o}{.}\PY{n}{logstd} \PY{o}{=} \PY{n}{logstd}
            \PY{n+nb+bp}{self}\PY{o}{.}\PY{n}{model}\PY{o}{.}\PY{n}{trainable\PYZus{}variables}\PY{o}{.}\PY{n}{append}\PY{p}{(}\PY{n}{logstd}\PY{p}{)}

    \PY{k}{def} \PY{n+nf}{\PYZus{}build\PYZus{}net}\PY{p}{(}\PY{n+nb+bp}{self}\PY{p}{)}\PY{p}{:}
        \PY{k}{pass}

    \PY{k}{def} \PY{n+nf+fm}{\PYZus{}\PYZus{}call\PYZus{}\PYZus{}}\PY{p}{(}\PY{n+nb+bp}{self}\PY{p}{,} \PY{n}{state}\PY{p}{)}\PY{p}{:}
        \PY{k}{if} \PY{n+nb+bp}{self}\PY{o}{.}\PY{n}{flag} \PY{o}{==} \PY{l+s+s1}{\PYZsq{}}\PY{l+s+s1}{ddpg}\PY{l+s+s1}{\PYZsq{}}\PY{p}{:}
            \PY{k}{return} \PY{n}{tf}\PY{o}{.}\PY{n}{squeeze}\PY{p}{(}\PY{n+nb+bp}{self}\PY{o}{.}\PY{n}{model}\PY{p}{(}\PY{n}{state}\PY{p}{)}\PY{p}{)}
        \PY{k}{elif} \PY{n+nb+bp}{self}\PY{o}{.}\PY{n}{flag} \PY{o}{==} \PY{l+s+s1}{\PYZsq{}}\PY{l+s+s1}{ppo}\PY{l+s+s1}{\PYZsq{}}\PY{p}{:} 
            \PY{n}{mean} \PY{o}{=} \PY{n}{tf}\PY{o}{.}\PY{n}{squeeze}\PY{p}{(}\PY{n+nb+bp}{self}\PY{o}{.}\PY{n}{model}\PY{p}{(}\PY{n}{state}\PY{p}{)}\PY{p}{)}
            \PY{n}{std} \PY{o}{=} \PY{n}{tf}\PY{o}{.}\PY{n}{squeeze}\PY{p}{(}\PY{n}{tf}\PY{o}{.}\PY{n}{exp}\PY{p}{(}\PY{n+nb+bp}{self}\PY{o}{.}\PY{n}{model}\PY{o}{.}\PY{n}{logstd}\PY{p}{)}\PY{p}{)}
            \PY{k}{return} \PY{n}{mean}\PY{p}{,} \PY{n}{std}

    \PY{k}{def} \PY{n+nf}{train}\PY{p}{(}\PY{n+nb+bp}{self}\PY{p}{)}\PY{p}{:}
        \PY{k}{if} \PY{n+nb+bp}{self}\PY{o}{.}\PY{n}{flag} \PY{o}{==} \PY{l+s+s1}{\PYZsq{}}\PY{l+s+s1}{ddpg}\PY{l+s+s1}{\PYZsq{}}\PY{p}{:}
            \PY{n+nb+bp}{self}\PY{o}{.}\PY{n}{ddpg\PYZus{}train}\PY{p}{(}\PY{p}{)}
        \PY{k}{elif} \PY{n+nb+bp}{self}\PY{o}{.}\PY{n}{flag} \PY{o}{==} \PY{l+s+s1}{\PYZsq{}}\PY{l+s+s1}{ppo}\PY{l+s+s1}{\PYZsq{}}\PY{p}{:}
            \PY{n+nb+bp}{self}\PY{o}{.}\PY{n}{ppo\PYZus{}train}\PY{p}{(}\PY{p}{)}

    \PY{k}{def} \PY{n+nf}{update\PYZus{}target}\PY{p}{(}\PY{n+nb+bp}{self}\PY{p}{)}\PY{p}{:}
        \PY{k}{pass}
\end{Verbatim}

%% file: pygtex_files/critic-py.tex
\begin{Verbatim}[commandchars=\\\{\},baselinestretch=0.9,frame=leftline,framesep=1.5ex,framerule=0.8pt,fontsize=\tiny]
\PY{k}{class} \PY{n+nc}{Critic}\PY{p}{:}
    \PY{k}{def} \PY{n+nf+fm}{\PYZus{}\PYZus{}init\PYZus{}\PYZus{}}\PY{p}{(}\PY{n+nb+bp}{self}\PY{p}{,} \PY{n}{state\PYZus{}size}\PY{p}{,} \PY{n}{action\PYZus{}size}\PY{p}{)}\PY{p}{:}
        \PY{n+nb+bp}{self}\PY{o}{.}\PY{n}{model} \PY{o}{=} \PY{n}{\PYZus{}build\PYZus{}net}\PY{p}{(}\PY{p}{)}
        \PY{n+nb+bp}{self}\PY{o}{.}\PY{n}{optim} \PY{o}{=} \PY{n}{tf}\PY{o}{.}\PY{n}{keras}\PY{o}{.}\PY{n}{optimizers}\PY{o}{.}\PY{n}{Adam}\PY{p}{(}\PY{n+nb+bp}{self}\PY{o}{.}\PY{n}{lr}\PY{p}{)}
        \PY{k}{if} \PY{n}{flag} \PY{o}{==} \PY{l+s+s1}{\PYZsq{}}\PY{l+s+s1}{ddpg}\PY{l+s+s1}{\PYZsq{}}\PY{p}{:}
            \PY{n+nb+bp}{self}\PY{o}{.}\PY{n}{target} \PY{o}{=} \PY{n}{\PYZus{}build\PYZus{}net}\PY{p}{(}\PY{p}{)}
            \PY{n+nb+bp}{self}\PY{o}{.}\PY{n}{target}\PY{o}{.}\PY{n}{set\PYZus{}weights}\PY{p}{(}\PY{n+nb+bp}{self}\PY{o}{.}\PY{n}{model}\PY{o}{.}\PY{n}{get\PYZus{}weights}\PY{p}{(}\PY{p}{)}\PY{p}{)}

    \PY{k}{def} \PY{n+nf+fm}{\PYZus{}\PYZus{}call\PYZus{}\PYZus{}}\PY{p}{(}\PY{n+nb+bp}{self}\PY{p}{,} \PY{n}{state}\PY{p}{)}\PY{p}{:}
        \PY{k}{return} \PY{n}{tf}\PY{o}{.}\PY{n}{squeeze}\PY{p}{(}\PY{n+nb+bp}{self}\PY{o}{.}\PY{n}{model}\PY{p}{(}\PY{n}{state}\PY{p}{)}\PY{p}{)}

    \PY{k}{def} \PY{n+nf}{\PYZus{}build\PYZus{}net}\PY{p}{(}\PY{n+nb+bp}{self}\PY{p}{)}\PY{p}{:}
        \PY{k}{pass}

    \PY{k}{def} \PY{n+nf}{train}\PY{p}{(}\PY{n+nb+bp}{self}\PY{p}{)}\PY{p}{:}
        \PY{k}{if} \PY{n+nb+bp}{self}\PY{o}{.}\PY{n}{flag} \PY{o}{==} \PY{l+s+s1}{\PYZsq{}}\PY{l+s+s1}{ddpg}\PY{l+s+s1}{\PYZsq{}}\PY{p}{:}
            \PY{n+nb+bp}{self}\PY{o}{.}\PY{n}{ddpg\PYZus{}train}\PY{p}{(}\PY{p}{)}
        \PY{k}{elif} \PY{n+nb+bp}{self}\PY{o}{.}\PY{n}{flag} \PY{o}{==} \PY{l+s+s1}{\PYZsq{}}\PY{l+s+s1}{ppo}\PY{l+s+s1}{\PYZsq{}}\PY{p}{:}
            \PY{n+nb+bp}{self}\PY{o}{.}\PY{n}{ppo\PYZus{}train}\PY{p}{(}\PY{p}{)}

    \PY{k}{def} \PY{n+nf}{update\PYZus{}target}\PY{p}{(}\PY{n+nb+bp}{self}\PY{p}{)}\PY{p}{:}
        \PY{k}{pass}
\end{Verbatim}

%% file: pygtex_files/agent-py.tex
\begin{Verbatim}[commandchars=\\\{\},baselinestretch=0.9,frame=leftline,framesep=1.5ex,framerule=0.8pt,fontsize=\tiny]
\PY{k}{class} \PY{n+nc}{Agent}\PY{p}{:}
    \PY{k}{def} \PY{n+nf+fm}{\PYZus{}\PYZus{}init\PYZus{}\PYZus{}}\PY{p}{(}\PY{n+nb+bp}{self}\PY{p}{,} \PY{n}{state\PYZus{}size}\PY{p}{,} \PY{n}{action\PYZus{}size}\PY{p}{)}\PY{p}{:}
        \PY{n+nb+bp}{self}\PY{o}{.}\PY{n}{actor} \PY{o}{=} \PY{n}{Actor}\PY{p}{(}\PY{n}{state\PYZus{}size}\PY{p}{,} \PY{n}{action\PYZus{}size}\PY{p}{)}
        \PY{n+nb+bp}{self}\PY{o}{.}\PY{n}{critic} \PY{o}{=} \PY{n}{Critic}\PY{p}{(}\PY{n}{state\PYZus{}size}\PY{p}{,} \PY{n}{action\PYZus{}size}\PY{p}{)}
        \PY{n+nb+bp}{self}\PY{o}{.}\PY{n}{buffer} \PY{o}{=} \PY{n}{Buffer}\PY{p}{(}\PY{n}{max\PYZus{}size}\PY{p}{,} \PY{n}{tmax}\PY{o}{=}\PY{l+m+mi}{1000}\PY{p}{)}

    \PY{k}{def} \PY{n+nf}{policy}\PY{p}{(}\PY{n+nb+bp}{self}\PY{p}{)}\PY{p}{:}
        \PY{k}{if} \PY{n+nb+bp}{self}\PY{o}{.}\PY{n}{flag} \PY{o}{==} \PY{l+s+s1}{\PYZsq{}}\PY{l+s+s1}{ddpg}\PY{l+s+s1}{\PYZsq{}}\PY{p}{:}
            \PY{n+nb+bp}{self}\PY{o}{.}\PY{n}{ddpg\PYZus{}policy}\PY{p}{(}\PY{p}{)}
        \PY{k}{elif} \PY{n+nb+bp}{self}\PY{o}{.}\PY{n}{flag} \PY{o}{==} \PY{l+s+s1}{\PYZsq{}}\PY{l+s+s1}{ppo}\PY{l+s+s1}{\PYZsq{}}\PY{p}{:}
            \PY{n+nb+bp}{self}\PY{o}{.}\PY{n}{ppo\PYZus{}policy}\PY{p}{(}\PY{p}{)}

    \PY{k}{def} \PY{n+nf}{train}\PY{p}{(}\PY{n+nb+bp}{self}\PY{p}{)}\PY{p}{:}
        \PY{k}{if} \PY{n+nb+bp}{self}\PY{o}{.}\PY{n}{flag} \PY{o}{==} \PY{l+s+s1}{\PYZsq{}}\PY{l+s+s1}{ddpg}\PY{l+s+s1}{\PYZsq{}}\PY{p}{:}
            \PY{n+nb+bp}{self}\PY{o}{.}\PY{n}{ddpg\PYZus{}train}\PY{p}{(}\PY{p}{)}
            \PY{n+nb+bp}{self}\PY{o}{.}\PY{n}{update\PYZus{}target}\PY{p}{(}\PY{p}{)} \PY{c+c1}{\PYZsh{} update the target models}
        \PY{k}{elif} \PY{n+nb+bp}{self}\PY{o}{.}\PY{n}{flag} \PY{o}{==} \PY{l+s+s1}{\PYZsq{}}\PY{l+s+s1}{ppo}\PY{l+s+s1}{\PYZsq{}}\PY{p}{:}
            \PY{n+nb+bp}{self}\PY{o}{.}\PY{n}{ppo\PYZus{}train}\PY{p}{(}\PY{p}{)}
            \PY{n+nb+bp}{self}\PY{o}{.}\PY{n}{buffer}\PY{o}{.}\PY{n}{clear}\PY{p}{(}\PY{p}{)} \PY{c+c1}{\PYZsh{} clear the buffer}

    \PY{k}{def} \PY{n+nf}{update\PYZus{}target}\PY{p}{(}\PY{n+nb+bp}{self}\PY{p}{)}\PY{p}{:}
        \PY{k}{if} \PY{n+nb+bp}{self}\PY{o}{.}\PY{n}{flag} \PY{o}{==} \PY{l+s+s1}{\PYZsq{}}\PY{l+s+s1}{ddpg}\PY{l+s+s1}{\PYZsq{}}\PY{p}{:}
            \PY{n+nb+bp}{self}\PY{o}{.}\PY{n}{actor}\PY{o}{.}\PY{n}{update\PYZus{}target}\PY{p}{(}\PY{p}{)}
            \PY{n+nb+bp}{self}\PY{o}{.}\PY{n}{critic}\PY{o}{.}\PY{n}{update\PYZus{}target}\PY{p}{(}\PY{p}{)}
\end{Verbatim}

%% file: pygtex_files/gae-py.tex
\begin{Verbatim}[commandchars=\\\{\},baselinestretch=0.9,frame=leftline,framesep=1.5ex,framerule=0.8pt,fontsize=\tiny]
\PY{k}{class} \PY{n+nc}{Agent}\PY{p}{:}
    \PY{k}{def} \PY{n+nf}{compute\PYZus{}advantages}\PY{p}{(}\PY{n+nb+bp}{self}\PY{p}{,} \PY{n}{states}\PY{p}{,} \PY{n}{next\PYZus{}states}\PY{p}{,} \PY{n}{rewards}\PY{p}{,} \PY{n}{dones}\PY{p}{)}\PY{p}{:}
        \PY{c+c1}{\PYZsh{} inputs are tensors and outputs are numpy arrays}
        \PY{n}{s\PYZus{}values} \PY{o}{=} \PY{n+nb+bp}{self}\PY{o}{.}\PY{n}{critic}\PY{p}{(}\PY{n}{states}\PY{p}{)}
        \PY{n}{ns\PYZus{}values} \PY{o}{=} \PY{n+nb+bp}{self}\PY{o}{.}\PY{n}{critic}\PY{p}{(}\PY{n}{next\PYZus{}states}\PY{p}{)}

        \PY{n}{adv} \PY{o}{=} \PY{n}{np}\PY{o}{.}\PY{n}{zeros}\PY{p}{(}\PY{n}{shape}\PY{o}{=}\PY{p}{(}\PY{n+nb}{len}\PY{p}{(}\PY{n}{rewards}\PY{p}{)}\PY{p}{,} \PY{p}{)}\PY{p}{)}
        \PY{n}{returns} \PY{o}{=} \PY{n}{np}\PY{o}{.}\PY{n}{zeros}\PY{p}{(}\PY{n}{shape}\PY{o}{=}\PY{p}{(}\PY{n+nb}{len}\PY{p}{(}\PY{n}{rewards}\PY{p}{)}\PY{p}{,} \PY{p}{)}\PY{p}{)}

        \PY{n}{discount} \PY{o}{=} \PY{n+nb+bp}{self}\PY{o}{.}\PY{n}{gamma}
        \PY{n}{lmbda} \PY{o}{=} \PY{n+nb+bp}{self}\PY{o}{.}\PY{n}{lmbda}
        \PY{n}{g} \PY{o}{=} \PY{l+m+mi}{0}
        \PY{n}{returns\PYZus{}current} \PY{o}{=} \PY{n}{ns\PYZus{}values}\PY{p}{[}\PY{o}{\PYZhy{}}\PY{l+m+mi}{1}\PY{p}{]}
        \PY{k}{for} \PY{n}{i} \PY{o+ow}{in} \PY{n+nb}{reversed}\PY{p}{(}\PY{n+nb}{range}\PY{p}{(}\PY{n+nb}{len}\PY{p}{(}\PY{n}{rewards}\PY{p}{)}\PY{p}{)}\PY{p}{)}\PY{p}{:}
            \PY{n}{gamma} \PY{o}{=} \PY{n}{discount} \PY{o}{*} \PY{p}{(}\PY{l+m+mf}{1.} \PY{o}{\PYZhy{}} \PY{n}{dones}\PY{p}{[}\PY{n}{i}\PY{p}{]}\PY{p}{)}
            \PY{n}{td\PYZus{}error} \PY{o}{=} \PY{n}{rewards}\PY{p}{[}\PY{n}{i}\PY{p}{]} \PY{o}{+} \PY{n}{gamma} \PY{o}{*} \PY{n}{ns\PYZus{}values}\PY{p}{[}\PY{n}{i}\PY{p}{]} \PY{o}{\PYZhy{}} \PY{n}{s\PYZus{}values}\PY{p}{[}\PY{n}{i}\PY{p}{]}
            \PY{n}{g} \PY{o}{=} \PY{n}{td\PYZus{}error} \PY{o}{+} \PY{n}{gamma} \PY{o}{*} \PY{n}{lmbda} \PY{o}{*} \PY{n}{g}
            \PY{n}{returns\PYZus{}current} \PY{o}{=} \PY{n}{rewards}\PY{p}{[}\PY{n}{i}\PY{p}{]} \PY{o}{+} \PY{n}{gamma} \PY{o}{*} \PY{n}{returns\PYZus{}current}
            \PY{n}{adv}\PY{p}{[}\PY{n}{i}\PY{p}{]} \PY{o}{=} \PY{n}{g}
            \PY{n}{returns}\PY{p}{[}\PY{n}{i}\PY{p}{]} \PY{o}{=} \PY{n}{returns\PYZus{}current}
        \PY{n}{adv} \PY{o}{=} \PY{p}{(}\PY{n}{adv} \PY{o}{\PYZhy{}} \PY{n}{np}\PY{o}{.}\PY{n}{mean}\PY{p}{(}\PY{n}{adv}\PY{p}{)}\PY{p}{)} \PY{o}{/} \PY{p}{(}\PY{n}{np}\PY{o}{.}\PY{n}{std}\PY{p}{(}\PY{n}{adv}\PY{p}{)} \PY{o}{+} \PY{l+m+mf}{1e\PYZhy{}10}\PY{p}{)}
        \PY{k}{return} \PY{n}{returns}\PY{p}{,} \PY{n}{adv}
\end{Verbatim}

%% file: pygtex_files/ppo_train-py.tex
\begin{Verbatim}[commandchars=\\\{\},baselinestretch=0.9,frame=leftline,framesep=1.5ex,framerule=0.8pt,fontsize=\tiny]
\PY{k+kn}{import} \PY{n+nn}{tensorflow\PYZus{}probability} \PY{k+kn}{as} \PY{n+nn}{tfp}
\PY{k}{class} \PY{n+nc}{Actor}\PY{p}{:}
    \PY{k}{def} \PY{n+nf}{ppo\PYZus{}train}\PY{p}{(}\PY{n+nb+bp}{self}\PY{p}{,} \PY{n}{state\PYZus{}batch}\PY{p}{,} \PY{n}{action\PYZus{}batch}\PY{p}{,} \PY{n}{advantages}\PY{p}{,} \PY{n}{old\PYZus{}pi}\PY{p}{)}\PY{p}{:}
        \PY{k}{with} \PY{n}{tf}\PY{o}{.}\PY{n}{GradientTape}\PY{p}{(}\PY{p}{)} \PY{k}{as} \PY{n}{tape}\PY{p}{:}
            \PY{n}{mean} \PY{o}{=} \PY{n}{tf}\PY{o}{.}\PY{n}{squeeze}\PY{p}{(}\PY{n+nb+bp}{self}\PY{o}{.}\PY{n}{model}\PY{p}{(}\PY{n}{state\PYZus{}batch}\PY{p}{)}\PY{p}{)}
            \PY{n}{std} \PY{o}{=} \PY{n}{tf}\PY{o}{.}\PY{n}{squeeze}\PY{p}{(}\PY{n}{tf}\PY{o}{.}\PY{n}{exp}\PY{p}{(}\PY{n+nb+bp}{self}\PY{o}{.}\PY{n}{model}\PY{o}{.}\PY{n}{logstd}\PY{p}{)}\PY{p}{)}
            \PY{n}{pi} \PY{o}{=} \PY{n}{tfp}\PY{o}{.}\PY{n}{distributions}\PY{o}{.}\PY{n}{Normal}\PY{p}{(}\PY{n}{mean}\PY{p}{,} \PY{n}{std}\PY{p}{)}
            \PY{n}{ratio} \PY{o}{=} \PY{n}{tf}\PY{o}{.}\PY{n}{exp}\PY{p}{(}\PY{n}{pi}\PY{o}{.}\PY{n}{log\PYZus{}prob}\PY{p}{(}\PY{n}{tf}\PY{o}{.}\PY{n}{squeeze}\PY{p}{(}\PY{n}{action\PYZus{}batch}\PY{p}{)}\PY{p}{)} \PY{o}{\PYZhy{}}
                           \PY{n}{old\PYZus{}pi}\PY{o}{.}\PY{n}{log\PYZus{}prob}\PY{p}{(}\PY{n}{tf}\PY{o}{.}\PY{n}{squeeze}\PY{p}{(}\PY{n}{action\PYZus{}batch}\PY{p}{)}\PY{p}{)}\PY{p}{)}
            \PY{n}{surr} \PY{o}{=} \PY{n}{ratio} \PY{o}{*} \PY{n}{advantages}  \PY{c+c1}{\PYZsh{} surrogate function}
            \PY{n}{kl} \PY{o}{=} \PY{n}{tfp}\PY{o}{.}\PY{n}{distributions}\PY{o}{.}\PY{n}{kl\PYZus{}divergence}\PY{p}{(}\PY{n}{old\PYZus{}pi}\PY{p}{,} \PY{n}{pi}\PY{p}{)}
            \PY{n+nb+bp}{self}\PY{o}{.}\PY{n}{kl\PYZus{}value} \PY{o}{=} \PY{n}{tf}\PY{o}{.}\PY{n}{reduce\PYZus{}mean}\PY{p}{(}\PY{n}{kl}\PY{p}{)}
            \PY{k}{if} \PY{n+nb+bp}{self}\PY{o}{.}\PY{n}{method} \PY{o}{==} \PY{l+s+s1}{\PYZsq{}}\PY{l+s+s1}{penalty}\PY{l+s+s1}{\PYZsq{}}\PY{p}{:}  \PY{c+c1}{\PYZsh{} ppo\PYZhy{}penalty method}
                \PY{n}{actor\PYZus{}loss} \PY{o}{=} \PY{o}{\PYZhy{}}\PY{p}{(}\PY{n}{tf}\PY{o}{.}\PY{n}{reduce\PYZus{}mean}\PY{p}{(}\PY{n}{surr} \PY{o}{\PYZhy{}} \PY{n+nb+bp}{self}\PY{o}{.}\PY{n}{lam} \PY{o}{*} \PY{n}{kl}\PY{p}{)}\PY{p}{)}
                \PY{c+c1}{\PYZsh{} self.update\PYZus{}lambda()}
            \PY{k}{elif} \PY{n+nb+bp}{self}\PY{o}{.}\PY{n}{method} \PY{o}{==} \PY{l+s+s1}{\PYZsq{}}\PY{l+s+s1}{clip}\PY{l+s+s1}{\PYZsq{}}\PY{p}{:}  \PY{c+c1}{\PYZsh{} ppo\PYZhy{}clip method}
                \PY{n}{actor\PYZus{}loss} \PY{o}{=} \PY{o}{\PYZhy{}} \PY{n}{tf}\PY{o}{.}\PY{n}{reduce\PYZus{}mean}\PY{p}{(}
                    \PY{n}{tf}\PY{o}{.}\PY{n}{minimum}\PY{p}{(}\PY{n}{surr}\PY{p}{,} \PY{n}{tf}\PY{o}{.}\PY{n}{clip\PYZus{}by\PYZus{}value}\PY{p}{(}\PY{n}{ratio}\PY{p}{,}\PYZbs{}
                            \PY{l+m+mf}{1.} \PY{o}{\PYZhy{}} \PY{n+nb+bp}{self}\PY{o}{.}\PY{n}{epsilon}\PY{p}{,} \PY{l+m+mf}{1.} \PY{o}{+} \PY{n+nb+bp}{self}\PY{o}{.}\PY{n}{epsilon}\PY{p}{)} \PY{o}{*} \PY{n}{advantages}\PY{p}{)}\PY{p}{)}
            \PY{n}{actor\PYZus{}weights} \PY{o}{=} \PY{n+nb+bp}{self}\PY{o}{.}\PY{n}{model}\PY{o}{.}\PY{n}{trainable\PYZus{}variables}

        \PY{c+c1}{\PYZsh{} outside gradient tape}
        \PY{n}{actor\PYZus{}grad} \PY{o}{=} \PY{n}{tape}\PY{o}{.}\PY{n}{gradient}\PY{p}{(}\PY{n}{actor\PYZus{}loss}\PY{p}{,} \PY{n}{actor\PYZus{}weights}\PY{p}{)}
        \PY{n+nb+bp}{self}\PY{o}{.}\PY{n}{optimizer}\PY{o}{.}\PY{n}{apply\PYZus{}gradients}\PY{p}{(}\PY{n+nb}{zip}\PY{p}{(}\PY{n}{actor\PYZus{}grad}\PY{p}{,} \PY{n}{actor\PYZus{}weights}\PY{p}{)}\PY{p}{)}

        \PY{k}{return} \PY{n}{actor\PYZus{}loss}\PY{o}{.}\PY{n}{numpy}\PY{p}{(}\PY{p}{)}\PY{p}{,} \PY{n+nb+bp}{self}\PY{o}{.}\PY{n}{kl\PYZus{}value}\PY{o}{.}\PY{n}{numpy}\PY{p}{(}\PY{p}{)}

\PY{k}{class} \PY{n+nc}{Critic}\PY{p}{:}
     \PY{k}{def} \PY{n+nf}{ppo\PYZus{}train}\PY{p}{(}\PY{n+nb+bp}{self}\PY{p}{,} \PY{n}{state\PYZus{}batch}\PY{p}{,} \PY{n}{disc\PYZus{}rewards}\PY{p}{)}\PY{p}{:}
        \PY{k}{with} \PY{n}{tf}\PY{o}{.}\PY{n}{GradientTape}\PY{p}{(}\PY{p}{)} \PY{k}{as} \PY{n}{tape}\PY{p}{:}
            \PY{n}{critic\PYZus{}weights} \PY{o}{=} \PY{n+nb+bp}{self}\PY{o}{.}\PY{n}{model}\PY{o}{.}\PY{n}{trainable\PYZus{}variables}
            \PY{n}{critic\PYZus{}value} \PY{o}{=} \PY{n}{tf}\PY{o}{.}\PY{n}{squeeze}\PY{p}{(}\PY{n+nb+bp}{self}\PY{o}{.}\PY{n}{model}\PY{p}{(}\PY{n}{state\PYZus{}batch}\PY{p}{)}\PY{p}{)}
            \PY{n}{critic\PYZus{}loss} \PY{o}{=} \PY{n}{tf}\PY{o}{.}\PY{n}{math}\PY{o}{.}\PY{n}{reduce\PYZus{}mean}\PY{p}{(}\PY{n}{tf}\PY{o}{.}\PY{n}{square}\PY{p}{(}\PY{n}{disc\PYZus{}rewards} \PY{o}{\PYZhy{}} \PY{n}{critic\PYZus{}value}\PY{p}{)}\PY{p}{)}
        \PY{n}{critic\PYZus{}grad} \PY{o}{=} \PY{n}{tape}\PY{o}{.}\PY{n}{gradient}\PY{p}{(}\PY{n}{critic\PYZus{}loss}\PY{p}{,} \PY{n}{critic\PYZus{}weights}\PY{p}{)}
        \PY{n+nb+bp}{self}\PY{o}{.}\PY{n}{optimizer}\PY{o}{.}\PY{n}{apply\PYZus{}gradients}\PY{p}{(}\PY{n+nb}{zip}\PY{p}{(}\PY{n}{critic\PYZus{}grad}\PY{p}{,} \PY{n}{critic\PYZus{}weights}\PY{p}{)}\PY{p}{)}
\end{Verbatim}

%% file: pygtex_files/ppo_agent_train-py.tex
\begin{Verbatim}[commandchars=\\\{\},baselinestretch=0.9,frame=leftline,framesep=1.5ex,framerule=0.8pt,fontsize=\tiny]
\PY{k}{class} \PY{n+nc}{Agent}\PY{p}{:}
    \PY{k}{def} \PY{n+nf}{ppo\PYZus{}train}\PY{p}{(}\PY{n+nb+bp}{self}\PY{p}{,} \PY{n}{training\PYZus{}epochs}\PY{o}{=}\PY{l+m+mi}{20}\PY{p}{,} \PY{n}{tmax}\PY{o}{=}\PY{l+m+mi}{10000}\PY{p}{)}\PY{p}{:}
        \PY{c+c1}{\PYZsh{} make sure to have enough samples in buffer for training}
        \PY{k}{if} \PY{n}{tmax} \PY{o+ow}{is} \PY{o+ow}{not} \PY{n+nb+bp}{None} \PY{o+ow}{and} \PY{n+nb}{len}\PY{p}{(}\PY{n+nb+bp}{self}\PY{o}{.}\PY{n}{buffer}\PY{p}{)} \PY{o}{\PYZlt{}} \PY{n}{tmax}\PY{p}{:}
            \PY{k}{return} \PY{l+m+mi}{0}\PY{p}{,} \PY{l+m+mi}{0}\PY{p}{,} \PY{l+m+mi}{0}

        \PY{n}{n\PYZus{}split} \PY{o}{=} \PY{n+nb}{len}\PY{p}{(}\PY{n+nb+bp}{self}\PY{o}{.}\PY{n}{buffer}\PY{p}{)} \PY{o}{/}\PY{o}{/} \PY{n+nb+bp}{self}\PY{o}{.}\PY{n}{batch\PYZus{}size}
        \PY{n}{n\PYZus{}samples} \PY{o}{=} \PY{n}{n\PYZus{}split} \PY{o}{*} \PY{n+nb+bp}{self}\PY{o}{.}\PY{n}{batch\PYZus{}size}

        \PY{n}{s\PYZus{}batch}\PY{p}{,} \PY{n}{a\PYZus{}batch}\PY{p}{,} \PY{n}{r\PYZus{}batch}\PY{p}{,} \PY{n}{ns\PYZus{}batch}\PY{p}{,} \PY{n}{d\PYZus{}batch} \PY{o}{=} \PYZbs{}
            \PY{n+nb+bp}{self}\PY{o}{.}\PY{n}{buffer}\PY{o}{.}\PY{n}{get\PYZus{}samples}\PY{p}{(}\PY{n}{n\PYZus{}samples}\PY{p}{)}

        \PY{n}{s\PYZus{}batch} \PY{o}{=} \PY{n}{tf}\PY{o}{.}\PY{n}{convert\PYZus{}to\PYZus{}tensor}\PY{p}{(}\PY{n}{s\PYZus{}batch}\PY{p}{,} \PY{n}{dtype}\PY{o}{=}\PY{n}{tf}\PY{o}{.}\PY{n}{float32}\PY{p}{)}
        \PY{n}{a\PYZus{}batch} \PY{o}{=} \PY{n}{tf}\PY{o}{.}\PY{n}{convert\PYZus{}to\PYZus{}tensor}\PY{p}{(}\PY{n}{a\PYZus{}batch}\PY{p}{,} \PY{n}{dtype}\PY{o}{=}\PY{n}{tf}\PY{o}{.}\PY{n}{float32}\PY{p}{)}
        \PY{n}{r\PYZus{}batch} \PY{o}{=} \PY{n}{tf}\PY{o}{.}\PY{n}{convert\PYZus{}to\PYZus{}tensor}\PY{p}{(}\PY{n}{r\PYZus{}batch}\PY{p}{,} \PY{n}{dtype}\PY{o}{=}\PY{n}{tf}\PY{o}{.}\PY{n}{float32}\PY{p}{)}
        \PY{n}{ns\PYZus{}batch} \PY{o}{=} \PY{n}{tf}\PY{o}{.}\PY{n}{convert\PYZus{}to\PYZus{}tensor}\PY{p}{(}\PY{n}{ns\PYZus{}batch}\PY{p}{,} \PY{n}{dtype}\PY{o}{=}\PY{n}{tf}\PY{o}{.}\PY{n}{float32}\PY{p}{)}
        \PY{n}{d\PYZus{}batch} \PY{o}{=} \PY{n}{tf}\PY{o}{.}\PY{n}{convert\PYZus{}to\PYZus{}tensor}\PY{p}{(}\PY{n}{d\PYZus{}batch}\PY{p}{,} \PY{n}{dtype}\PY{o}{=}\PY{n}{tf}\PY{o}{.}\PY{n}{float32}\PY{p}{)}

        \PY{c+c1}{\PYZsh{} compute GAE and discounted returns}
        \PY{n}{target\PYZus{}values}\PY{p}{,} \PY{n}{advantages} \PY{o}{=} \PY{n+nb+bp}{self}\PY{o}{.}\PY{n}{compute\PYZus{}advantages}\PY{p}{(}\PY{n}{s\PYZus{}batch}\PY{p}{,} \PYZbs{}
                                                \PY{n}{ns\PYZus{}batch}\PY{p}{,} \PY{n}{r\PYZus{}batch}\PY{p}{,} \PY{n}{d\PYZus{}batch}\PY{p}{)}

        \PY{n}{advantages} \PY{o}{=} \PY{n}{tf}\PY{o}{.}\PY{n}{convert\PYZus{}to\PYZus{}tensor}\PY{p}{(}\PY{n}{advantages}\PY{p}{,} \PY{n}{dtype}\PY{o}{=}\PY{n}{tf}\PY{o}{.}\PY{n}{float32}\PY{p}{)}
        \PY{n}{disc\PYZus{}sum\PYZus{}reward} \PY{o}{=} \PY{n}{tf}\PY{o}{.}\PY{n}{convert\PYZus{}to\PYZus{}tensor}\PY{p}{(}
                \PY{n}{disc\PYZus{}sum\PYZus{}reward}\PY{p}{,} \PY{n}{dtype}\PY{o}{=}\PY{n}{tf}\PY{o}{.}\PY{n}{float32}\PY{p}{)}

        \PY{c+c1}{\PYZsh{} current policy}
        \PY{n}{mean}\PY{p}{,} \PY{n}{std} \PY{o}{=} \PY{n+nb+bp}{self}\PY{o}{.}\PY{n}{actor}\PY{p}{(}\PY{n}{s\PYZus{}batch}\PY{p}{)}
        \PY{n}{pi} \PY{o}{=} \PY{n}{tfp}\PY{o}{.}\PY{n}{distributions}\PY{o}{.}\PY{n}{Normal}\PY{p}{(}\PY{n}{mean}\PY{p}{,} \PY{n}{std}\PY{p}{)}

        \PY{n}{s\PYZus{}split} \PY{o}{=} \PY{n}{tf}\PY{o}{.}\PY{n}{split}\PY{p}{(}\PY{n}{s\PYZus{}batch}\PY{p}{,} \PY{n}{n\PYZus{}split}\PY{p}{)}
        \PY{n}{a\PYZus{}split} \PY{o}{=} \PY{n}{tf}\PY{o}{.}\PY{n}{split}\PY{p}{(}\PY{n}{a\PYZus{}batch}\PY{p}{,} \PY{n}{n\PYZus{}split}\PY{p}{)}
        \PY{n}{dr\PYZus{}split} \PY{o}{=} \PY{n}{tf}\PY{o}{.}\PY{n}{split}\PY{p}{(}\PY{n}{disc\PYZus{}sum\PYZus{}reward}\PY{p}{,} \PY{n}{n\PYZus{}split}\PY{p}{)}
        \PY{n}{adv\PYZus{}split} \PY{o}{=} \PY{n}{tf}\PY{o}{.}\PY{n}{split}\PY{p}{(}\PY{n}{advantages}\PY{p}{,} \PY{n}{n\PYZus{}split}\PY{p}{)}
        \PY{n}{indexes} \PY{o}{=} \PY{n}{np}\PY{o}{.}\PY{n}{arange}\PY{p}{(}\PY{n}{n\PYZus{}split}\PY{p}{,} \PY{n}{dtype}\PY{o}{=}\PY{n+nb}{int}\PY{p}{)}

        \PY{n}{a\PYZus{}loss\PYZus{}list} \PY{o}{=} \PY{p}{[}\PY{p}{]}
        \PY{n}{c\PYZus{}loss\PYZus{}list} \PY{o}{=} \PY{p}{[}\PY{p}{]}
        \PY{n}{kld\PYZus{}list} \PY{o}{=} \PY{p}{[}\PY{p}{]}
        \PY{n}{np}\PY{o}{.}\PY{n}{random}\PY{o}{.}\PY{n}{shuffle}\PY{p}{(}\PY{n}{indexes}\PY{p}{)}
        \PY{k}{for} \PY{n}{\PYZus{}} \PY{o+ow}{in} \PY{n+nb}{range}\PY{p}{(}\PY{n}{training\PYZus{}epochs}\PY{p}{)}\PY{p}{:}
            \PY{k}{for} \PY{n}{i} \PY{o+ow}{in} \PY{n}{indexes}\PY{p}{:}
                \PY{n}{old\PYZus{}pi} \PY{o}{=} \PY{n}{pi}\PY{p}{[}\PY{n}{i}\PY{o}{*}\PY{n+nb+bp}{self}\PY{o}{.}\PY{n}{batch\PYZus{}size}\PY{p}{:} \PY{p}{(}\PY{n}{i}\PY{o}{+}\PY{l+m+mi}{1}\PY{p}{)}\PY{o}{*}\PY{n+nb+bp}{self}\PY{o}{.}\PY{n}{batch\PYZus{}size}\PY{p}{]}

                \PY{c+c1}{\PYZsh{} update actor}
                \PY{n}{a\PYZus{}loss}\PY{p}{,} \PY{n}{kld} \PY{o}{=} \PY{n+nb+bp}{self}\PY{o}{.}\PY{n}{actor}\PY{o}{.}\PY{n}{ppo\PYZus{}train}\PY{p}{(}\PY{n}{s\PYZus{}split}\PY{p}{[}\PY{n}{i}\PY{p}{]}\PY{p}{,} 
                        \PY{n}{a\PYZus{}split}\PY{p}{[}\PY{n}{i}\PY{p}{]}\PY{p}{,} \PY{n}{adv\PYZus{}split}\PY{p}{[}\PY{n}{i}\PY{p}{]}\PY{p}{,} \PY{n}{old\PYZus{}pi}\PY{p}{)}
                \PY{n}{a\PYZus{}loss\PYZus{}list}\PY{o}{.}\PY{n}{append}\PY{p}{(}\PY{n}{a\PYZus{}loss}\PY{p}{)}
                \PY{n}{kld\PYZus{}list}\PY{o}{.}\PY{n}{append}\PY{p}{(}\PY{n}{kld}\PY{p}{)}

                \PY{c+c1}{\PYZsh{} update critic}
                \PY{n}{c\PYZus{}loss\PYZus{}list}\PY{o}{.}\PY{n}{append}\PY{p}{(}
                        \PY{n+nb+bp}{self}\PY{o}{.}\PY{n}{critic}\PY{o}{.}\PY{n}{ppo\PYZus{}train}\PY{p}{(}\PY{n}{s\PYZus{}split}\PY{p}{[}\PY{n}{i}\PY{p}{]}\PY{p}{,} \PY{n}{dr\PYZus{}split}\PY{p}{[}\PY{n}{i}\PY{p}{]}\PY{p}{)}\PY{p}{)}

            \PY{c+c1}{\PYZsh{} update lambda after each epoch}
            \PY{k}{if} \PY{n+nb+bp}{self}\PY{o}{.}\PY{n}{method} \PY{o}{==} \PY{l+s+s1}{\PYZsq{}}\PY{l+s+s1}{penalty}\PY{l+s+s1}{\PYZsq{}}\PY{p}{:}
                \PY{n+nb+bp}{self}\PY{o}{.}\PY{n}{actor}\PY{o}{.}\PY{n}{update\PYZus{}beta}\PY{p}{(}\PY{p}{)}

        \PY{n}{actor\PYZus{}loss} \PY{o}{=} \PY{n}{np}\PY{o}{.}\PY{n}{mean}\PY{p}{(}\PY{n}{a\PYZus{}loss\PYZus{}list}\PY{p}{)}
        \PY{n}{critic\PYZus{}loss} \PY{o}{=} \PY{n}{np}\PY{o}{.}\PY{n}{mean}\PY{p}{(}\PY{n}{c\PYZus{}loss\PYZus{}list}\PY{p}{)}
        \PY{n}{mean\PYZus{}kld} \PY{o}{=} \PY{n}{np}\PY{o}{.}\PY{n}{mean}\PY{p}{(}\PY{n}{kld\PYZus{}list}\PY{p}{)}

        \PY{c+c1}{\PYZsh{} clear the buffer  \PYZhy{}\PYZhy{} this is important}
        \PY{n+nb+bp}{self}\PY{o}{.}\PY{n}{buffer}\PY{o}{.}\PY{n}{clear}\PY{p}{(}\PY{p}{)}

        \PY{k}{return} \PY{n}{actor\PYZus{}loss}\PY{p}{,} \PY{n}{critic\PYZus{}loss}\PY{p}{,} \PY{n}{mean\PYZus{}kld}
\end{Verbatim}

%% file: pygtex_files/ddpg_train-py.tex
\begin{Verbatim}[commandchars=\\\{\},baselinestretch=0.9,frame=leftline,framesep=1.5ex,framerule=0.8pt,fontsize=\tiny]
\PY{k}{class} \PY{n+nc}{Actor}\PY{p}{:}
    \PY{k}{def} \PY{n+nf}{ddpg\PYZus{}train}\PY{p}{(}\PY{n+nb+bp}{self}\PY{p}{,} \PY{n}{state\PYZus{}batch}\PY{p}{,} \PY{n}{critic}\PY{p}{)}\PY{p}{:}
        \PY{k}{with} \PY{n}{tf}\PY{o}{.}\PY{n}{GradientTape}\PY{p}{(}\PY{p}{)} \PY{k}{as} \PY{n}{tape}\PY{p}{:}
            \PY{n}{actor\PYZus{}weights} \PY{o}{=} \PY{n+nb+bp}{self}\PY{o}{.}\PY{n}{model}\PY{o}{.}\PY{n}{trainable\PYZus{}variables}
            \PY{n}{actions} \PY{o}{=} \PY{n+nb+bp}{self}\PY{o}{.}\PY{n}{model}\PY{p}{(}\PY{n}{state\PYZus{}batch}\PY{p}{)}
            \PY{n}{critic\PYZus{}value} \PY{o}{=} \PY{n}{critic}\PY{o}{.}\PY{n}{model}\PY{p}{(}\PY{p}{[}\PY{n}{state\PYZus{}batch}\PY{p}{,} \PY{n}{actions}\PY{p}{]}\PY{p}{)}
            \PY{c+c1}{\PYZsh{} \PYZhy{}ve value is used to maximize value function}
            \PY{n}{actor\PYZus{}loss} \PY{o}{=} \PY{o}{\PYZhy{}}\PY{n}{tf}\PY{o}{.}\PY{n}{math}\PY{o}{.}\PY{n}{reduce\PYZus{}mean}\PY{p}{(}\PY{n}{critic\PYZus{}value}\PY{p}{)}
        \PY{n}{actor\PYZus{}grad} \PY{o}{=} \PY{n}{tape}\PY{o}{.}\PY{n}{gradient}\PY{p}{(}\PY{n}{actor\PYZus{}loss}\PY{p}{,} \PY{n}{actor\PYZus{}weights}\PY{p}{)}
        \PY{n+nb+bp}{self}\PY{o}{.}\PY{n}{optimizer}\PY{o}{.}\PY{n}{apply\PYZus{}gradients}\PY{p}{(}\PY{n+nb}{zip}\PY{p}{(}\PY{n}{actor\PYZus{}grad}\PY{p}{,} \PY{n}{actor\PYZus{}weights}\PY{p}{)}\PY{p}{)}

\PY{k}{class} \PY{n+nc}{Critic}\PY{p}{:}
    \PY{k}{def} \PY{n+nf}{ddpg\PYZus{}train}\PY{p}{(}\PY{n+nb+bp}{self}\PY{p}{,} \PY{n}{state\PYZus{}batch}\PY{p}{,} \PY{n}{action\PYZus{}batch}\PY{p}{,} \PY{n}{reward\PYZus{}batch}\PY{p}{,}
                                    \PY{n}{next\PYZus{}state\PYZus{}batch}\PY{p}{,} \PY{n}{actor}\PY{p}{)}\PY{p}{:}
        \PY{k}{with} \PY{n}{tf}\PY{o}{.}\PY{n}{GradientTape}\PY{p}{(}\PY{p}{)} \PY{k}{as} \PY{n}{tape}\PY{p}{:}
            \PY{n}{critic\PYZus{}weights} \PY{o}{=} \PY{n+nb+bp}{self}\PY{o}{.}\PY{n}{model}\PY{o}{.}\PY{n}{trainable\PYZus{}variables}
            \PY{n}{target\PYZus{}actions} \PY{o}{=} \PY{n}{actor}\PY{o}{.}\PY{n}{target}\PY{p}{(}\PY{n}{next\PYZus{}state\PYZus{}batch}\PY{p}{)}
            \PY{n}{target\PYZus{}critic} \PY{o}{=} \PY{n+nb+bp}{self}\PY{o}{.}\PY{n}{target}\PY{p}{(}\PY{p}{[}\PY{n}{next\PYZus{}state\PYZus{}batch}\PY{p}{,} \PY{n}{target\PYZus{}actions}\PY{p}{]}\PY{p}{)}
            \PY{n}{y} \PY{o}{=} \PY{n}{reward\PYZus{}batch} \PY{o}{+} \PY{n+nb+bp}{self}\PY{o}{.}\PY{n}{gamma} \PY{o}{*} \PY{n}{target\PYZus{}critic}
            \PY{n}{critic\PYZus{}value} \PY{o}{=} \PY{n+nb+bp}{self}\PY{o}{.}\PY{n}{model}\PY{p}{(}\PY{p}{[}\PY{n}{state\PYZus{}batch}\PY{p}{,} \PY{n}{action\PYZus{}batch}\PY{p}{]}\PY{p}{)}
            \PY{n}{critic\PYZus{}loss} \PY{o}{=} \PY{n}{tf}\PY{o}{.}\PY{n}{math}\PY{o}{.}\PY{n}{reduce\PYZus{}mean}\PY{p}{(}\PY{n}{tf}\PY{o}{.}\PY{n}{square}\PY{p}{(}\PY{n}{y} \PY{o}{\PYZhy{}} \PY{n}{critic\PYZus{}value}\PY{p}{)}\PY{p}{)}
        \PY{n}{critic\PYZus{}grad} \PY{o}{=} \PY{n}{tape}\PY{o}{.}\PY{n}{gradient}\PY{p}{(}\PY{n}{critic\PYZus{}loss}\PY{p}{,} \PY{n}{critic\PYZus{}weights}\PY{p}{)}
        \PY{n+nb+bp}{self}\PY{o}{.}\PY{n}{optimizer}\PY{o}{.}\PY{n}{apply\PYZus{}gradients}\PY{p}{(}\PY{n+nb}{zip}\PY{p}{(}\PY{n}{critic\PYZus{}grad}\PY{p}{,} \PY{n}{critic\PYZus{}weights}\PY{p}{)}\PY{p}{)}

\PY{k}{class} \PY{n+nc}{Agent}\PY{p}{:}
    \PY{k}{def} \PY{n+nf}{ddpg\PYZus{}train}\PY{p}{(}\PY{n+nb+bp}{self}\PY{p}{)}\PY{p}{:}
        \PY{c+c1}{\PYZsh{} sample from stored memory}
        \PY{n}{state\PYZus{}batch}\PY{p}{,} \PY{n}{action\PYZus{}batch}\PY{p}{,} \PY{n}{reward\PYZus{}batch}\PY{p}{,}\PYZbs{}
                        \PY{n}{next\PYZus{}state\PYZus{}batch} \PY{o}{=} \PY{n+nb+bp}{self}\PY{o}{.}\PY{n}{buffer}\PY{o}{.}\PY{n}{sample}\PY{p}{(}\PY{p}{)}
        \PY{c+c1}{\PYZsh{} train actor and critic models}
        \PY{n+nb+bp}{self}\PY{o}{.}\PY{n}{actor}\PY{o}{.}\PY{n}{ddpg\PYZus{}train}\PY{p}{(}\PY{n}{state\PYZus{}batch}\PY{p}{,} \PY{n+nb+bp}{self}\PY{o}{.}\PY{n}{critic}\PY{p}{)}
        \PY{n+nb+bp}{self}\PY{o}{.}\PY{n}{critic}\PY{o}{.}\PY{n}{ddpg\PYZus{}train}\PY{p}{(}\PY{n}{state\PYZus{}batch}\PY{p}{,} \PY{n}{action\PYZus{}batch}\PY{p}{,} \PY{n}{reward\PYZus{}batch}\PY{p}{,}
                          \PY{n}{next\PYZus{}state\PYZus{}batch}\PY{p}{,} \PY{n+nb+bp}{self}\PY{o}{.}\PY{n}{actor}\PY{p}{)}
\end{Verbatim}

%% file: pygtex_files/main-py.tex
\begin{Verbatim}[commandchars=\\\{\},baselinestretch=0.9,frame=leftline,framesep=1.5ex,framerule=0.8pt,fontsize=\tiny]
\PY{k+kn}{import} \PY{n+nn}{gym}
\PY{k}{if} \PY{n+nv+vm}{\PYZus{}\PYZus{}name\PYZus{}\PYZus{}} \PY{o}{==} \PY{l+s+s1}{\PYZsq{}}\PY{l+s+s1}{\PYZus{}\PYZus{}main\PYZus{}\PYZus{}}\PY{l+s+s1}{\PYZsq{}}\PY{p}{:}
    \PY{c+c1}{\PYZsh{} start open/AI GYM environment}
    \PY{n}{env} \PY{o}{=} \PY{n}{gym}\PY{o}{.}\PY{n}{make}\PY{p}{(}\PY{l+s+s1}{\PYZsq{}}\PY{l+s+s1}{Pendulum\PYZhy{}v0}\PY{l+s+s1}{\PYZsq{}}\PY{p}{)}
    \PY{c+c1}{\PYZsh{} Create an Actor\PYZhy{}Critic Agent}
    \PY{n}{agent} \PY{o}{=} \PY{n}{Agent}\PY{p}{(}\PY{n}{state\PYZus{}size}\PY{p}{,} \PY{n}{action\PYZus{}size}\PY{p}{,} \PY{n}{batch\PYZus{}size}\PY{o}{=}\PY{l+m+mi}{128}\PY{p}{,}
                         \PY{n}{upper\PYZus{}bound}\PY{o}{=}\PY{n}{upper\PYZus{}bound}\PY{p}{,} \PY{o}{.}\PY{o}{.}\PY{o}{.}\PY{p}{)}
    \PY{n}{state\PYZus{}size} \PY{o}{=} \PY{n}{env}\PY{o}{.}\PY{n}{observation\PYZus{}space}\PY{o}{.}\PY{n}{shape}
    \PY{n}{action\PYZus{}size} \PY{o}{=} \PY{n}{env}\PY{o}{.}\PY{n}{observation\PYZus{}space}\PY{o}{.}\PY{n}{shape}
    \PY{n}{upper\PYZus{}bound} \PY{o}{=} \PY{n}{env}\PY{o}{.}\PY{n}{action\PYZus{}space}\PY{o}{.}\PY{n}{high}\PY{p}{[}\PY{l+m+mi}{0}\PY{p}{]}
    \PY{c+c1}{\PYZsh{} training loop}
    \PY{n}{ep\PYZus{}reward\PYZus{}list} \PY{o}{=} \PY{p}{[}\PY{p}{]}
    \PY{k}{for} \PY{n}{ep} \PY{o+ow}{in} \PY{n+nb}{range}\PY{p}{(}\PY{n}{MAX\PYZus{}EPISODES}\PY{p}{)}\PY{p}{:}
        \PY{n}{state} \PY{o}{=} \PY{n}{env}\PY{o}{.}\PY{n}{reset}\PY{p}{(}\PY{p}{)}
        \PY{n}{ep\PYZus{}reward} \PY{o}{=} \PY{l+m+mi}{0}
        \PY{k}{while} \PY{n+nb+bp}{True}\PY{p}{:}
            \PY{c+c1}{\PYZsh{} take action based on current policy}
            \PY{n}{action} \PY{o}{=} \PY{n}{agent}\PY{o}{.}\PY{n}{policy}\PY{p}{(}\PY{n}{state}\PY{p}{)}
            \PY{c+c1}{\PYZsh{} gather rewards from the environment}
            \PY{n}{next\PYZus{}state}\PY{p}{,} \PY{n}{reward}\PY{p}{,} \PY{n}{done}\PY{p}{,} \PY{n}{\PYZus{}} \PY{o}{=} \PY{n}{env}\PY{o}{.}\PY{n}{step}\PY{p}{(}\PY{n}{action}\PY{p}{)}
            \PY{c+c1}{\PYZsh{} collect experiences}
            \PY{n}{agent}\PY{o}{.}\PY{n}{buffer}\PY{o}{.}\PY{n}{record}\PY{p}{(}\PY{n}{state}\PY{p}{,} \PY{n}{action}\PY{p}{,} \PY{n}{reward}\PY{p}{,} \PY{n}{next\PYZus{}state}\PY{p}{,} \PY{n}{done}\PY{p}{)}
            \PY{n}{ep\PYZus{}reward} \PY{o}{+}\PY{o}{=} \PY{n}{reward}
            \PY{c+c1}{\PYZsh{} train}
            \PY{n}{agent}\PY{o}{.}\PY{n}{train}\PY{p}{(}\PY{p}{)}
            \PY{n}{state} \PY{o}{=} \PY{n}{next\PYZus{}state}
            \PY{k}{if} \PY{n}{done}\PY{p}{:}
                \PY{n}{ep\PYZus{}reward\PYZus{}list}\PY{o}{.}\PY{n}{append}\PY{p}{(}\PY{n}{ep\PYZus{}reward}\PY{p}{)}
                \PY{k}{break}
        \PY{k}{if} \PY{n}{np}\PY{o}{.}\PY{n}{mean}\PY{p}{(}\PY{n}{ep\PYZus{}reward\PYZus{}list}\PY{p}{[}\PY{o}{\PYZhy{}}\PY{l+m+mi}{40}\PY{p}{:}\PY{p}{]}\PY{p}{)} \PY{o}{\PYZgt{}} \PY{o}{\PYZhy{}}\PY{l+m+mi}{200}\PY{p}{:}
            \PY{k}{print}\PY{p}{(}\PY{l+s+s1}{\PYZsq{}}\PY{l+s+s1}{Problem is solved in \PYZob{}\PYZcb{} episodes}\PY{l+s+s1}{\PYZsq{}}\PY{o}{.}\PY{n}{format}\PY{p}{(}\PY{n}{ep}\PY{p}{)}\PY{p}{)}
            \PY{k}{break}
    \PY{n}{env}\PY{o}{.}\PY{n}{close}\PY{p}{(}\PY{p}{)}
\end{Verbatim}

%% file: pygtex_files/policy-py.tex
\begin{Verbatim}[commandchars=\\\{\},baselinestretch=0.9,frame=leftline,framesep=1.5ex,framerule=0.8pt,fontsize=\tiny]
\PY{k}{class} \PY{n+nc}{Agent}\PY{p}{:}
    \PY{k}{def} \PY{n+nf}{ppo\PYZus{}policy}\PY{p}{(}\PY{n+nb+bp}{self}\PY{p}{,} \PY{n}{state}\PY{p}{)}\PY{p}{:}
        \PY{n}{tf\PYZus{}state} \PY{o}{=} \PY{n}{tf}\PY{o}{.}\PY{n}{expand\PYZus{}dims}\PY{p}{(}\PY{n}{tf}\PY{o}{.}\PY{n}{convert\PYZus{}to\PYZus{}tensor}\PY{p}{(}\PY{n}{state}\PY{p}{)}\PY{p}{,} \PY{l+m+mi}{0}\PY{p}{)}
        \PY{n}{mean}\PY{p}{,} \PY{n}{std} \PY{o}{=} \PY{n+nb+bp}{self}\PY{o}{.}\PY{n}{actor}\PY{p}{(}\PY{n}{tf\PYZus{}state}\PY{p}{)}
        \PY{n}{pi} \PY{o}{=} \PY{n}{tfp}\PY{o}{.}\PY{n}{distributions}\PY{o}{.}\PY{n}{Normal}\PY{p}{(}\PY{n}{mean}\PY{p}{,} \PY{n}{std}\PY{p}{)}
        \PY{c+c1}{\PYZsh{} sample from the current policy distribution}
        \PY{n}{action} \PY{o}{=} \PY{n}{pi}\PY{o}{.}\PY{n}{sample}\PY{p}{(}\PY{n}{sample\PYZus{}shape}\PY{o}{=}\PY{n+nb+bp}{self}\PY{o}{.}\PY{n}{action\PYZus{}size}\PY{p}{)}
        \PY{n}{valid\PYZus{}action} \PY{o}{=} \PY{n}{tf}\PY{o}{.}\PY{n}{clip\PYZus{}by\PYZus{}value}\PY{p}{(}\PY{n}{action}\PY{p}{,} 
                \PY{o}{\PYZhy{}}\PY{n+nb+bp}{self}\PY{o}{.}\PY{n}{upper\PYZus{}bound}\PY{p}{,} \PY{n+nb+bp}{self}\PY{o}{.}\PY{n}{upper\PYZus{}bound}\PY{p}{)}
        \PY{k}{return} \PY{n}{valid\PYZus{}action}\PY{o}{.}\PY{n}{numpy}\PY{p}{(}\PY{p}{)}

    \PY{k}{def} \PY{n+nf}{ddpg\PYZus{}policy}\PY{p}{(}\PY{n+nb+bp}{self}\PY{p}{,} \PY{n}{state}\PY{p}{)}\PY{p}{:}
        \PY{n}{tf\PYZus{}state} \PY{o}{=} \PY{n}{tf}\PY{o}{.}\PY{n}{expand\PYZus{}dims}\PY{p}{(}\PY{n}{tf}\PY{o}{.}\PY{n}{convert\PYZus{}to\PYZus{}tensor}\PY{p}{(}\PY{n}{state}\PY{p}{)}\PY{p}{,} \PY{l+m+mi}{0}\PY{p}{)}
        \PY{c+c1}{\PYZsh{} obtain action by using deterministic policy}
        \PY{n}{sampled\PYZus{}action} \PY{o}{=} \PY{n}{tf}\PY{o}{.}\PY{n}{squeeze}\PY{p}{(}\PY{n+nb+bp}{self}\PY{o}{.}\PY{n}{actor}\PY{o}{.}\PY{n}{model}\PY{p}{(}\PY{n}{tf\PYZus{}state}\PY{p}{)}\PY{p}{)}
        \PY{n}{noise} \PY{o}{=} \PY{n+nb+bp}{self}\PY{o}{.}\PY{n}{noise\PYZus{}object}\PY{p}{(}\PY{p}{)}
        \PY{c+c1}{\PYZsh{} Add noise to the action}
        \PY{n}{sampled\PYZus{}action} \PY{o}{=} \PY{n}{sampled\PYZus{}action}\PY{o}{.}\PY{n}{numpy}\PY{p}{(}\PY{p}{)} \PY{o}{+} \PY{n}{noise}
        \PY{c+c1}{\PYZsh{} Make sure that the action is within bounds}
        \PY{n}{valid\PYZus{}action} \PY{o}{=} \PY{n}{np}\PY{o}{.}\PY{n}{clip}\PY{p}{(}\PY{n}{sampled\PYZus{}action}\PY{p}{,} 
                \PY{n+nb+bp}{self}\PY{o}{.}\PY{n}{lower\PYZus{}bound}\PY{p}{,} \PY{n+nb+bp}{self}\PY{o}{.}\PY{n}{upper\PYZus{}bound}\PY{p}{)}
        \PY{k}{return} \PY{n}{valid\PYZus{}action}\PY{o}{.}\PY{n}{numpy}\PY{p}{(}\PY{p}{)}
\end{Verbatim}

%% file: pygtex_files/build_net-py.tex
\begin{Verbatim}[commandchars=\\\{\},baselinestretch=0.9,frame=leftline,framesep=1.5ex,framerule=0.8pt,fontsize=\tiny]
\PY{k+kn}{from} \PY{n+nn}{keras} \PY{k+kn}{import} \PY{n}{layers}
\PY{k}{class} \PY{n+nc}{Critic}\PY{p}{:}
    \PY{k}{def} \PY{n+nf}{\PYZus{}build\PYZus{}net}\PY{p}{(}\PY{n+nb+bp}{self}\PY{p}{)}\PY{p}{:}
        \PY{n}{state\PYZus{}input} \PY{o}{=} \PY{n}{layers}\PY{o}{.}\PY{n}{Input}\PY{p}{(}\PY{n}{shape}\PY{o}{=}\PY{n+nb+bp}{self}\PY{o}{.}\PY{n}{state\PYZus{}size}\PY{p}{)}
        \PY{n}{out} \PY{o}{=} \PY{n}{layers}\PY{o}{.}\PY{n}{Dense}\PY{p}{(}\PY{l+m+mi}{64}\PY{p}{,} \PY{n}{activation}\PY{o}{=}\PY{l+s+s2}{\PYZdq{}}\PY{l+s+s2}{relu}\PY{l+s+s2}{\PYZdq{}}\PY{p}{)}\PY{p}{(}\PY{n}{state\PYZus{}input}\PY{p}{)}
        \PY{n}{out} \PY{o}{=} \PY{n}{layers}\PY{o}{.}\PY{n}{Dense}\PY{p}{(}\PY{l+m+mi}{64}\PY{p}{,} \PY{n}{activation}\PY{o}{=}\PY{l+s+s2}{\PYZdq{}}\PY{l+s+s2}{relu}\PY{l+s+s2}{\PYZdq{}}\PY{p}{)}\PY{p}{(}\PY{n}{out}\PY{p}{)}
        \PY{n}{out} \PY{o}{=} \PY{n}{layers}\PY{o}{.}\PY{n}{Dense}\PY{p}{(}\PY{l+m+mi}{64}\PY{p}{,} \PY{n}{activation}\PY{o}{=}\PY{l+s+s2}{\PYZdq{}}\PY{l+s+s2}{relu}\PY{l+s+s2}{\PYZdq{}}\PY{p}{)}\PY{p}{(}\PY{n}{out}\PY{p}{)}
        \PY{n}{net\PYZus{}out} \PY{o}{=} \PY{n}{layers}\PY{o}{.}\PY{n}{Dense}\PY{p}{(}\PY{l+m+mi}{1}\PY{p}{)}\PY{p}{(}\PY{n}{out}\PY{p}{)}
        \PY{c+c1}{\PYZsh{} Outputs single value for give state\PYZhy{}action}
        \PY{n}{model} \PY{o}{=} \PY{n}{tf}\PY{o}{.}\PY{n}{keras}\PY{o}{.}\PY{n}{Model}\PY{p}{(}\PY{n}{inputs}\PY{o}{=}\PY{n}{state\PYZus{}input}\PY{p}{,} \PY{n}{outputs}\PY{o}{=}\PY{n}{net\PYZus{}out}\PY{p}{)}
        \PY{n}{model}\PY{o}{.}\PY{n}{summary}\PY{p}{(}\PY{p}{)}
        \PY{k}{return} \PY{n}{model}

\PY{k}{class} \PY{n+nc}{Actor}\PY{p}{:}
    \PY{k}{def} \PY{n+nf}{\PYZus{}build\PYZus{}net}\PY{p}{(}\PY{n+nb+bp}{self}\PY{p}{)}\PY{p}{:}
        \PY{n}{last\PYZus{}init} \PY{o}{=} \PY{n}{tf}\PY{o}{.}\PY{n}{random\PYZus{}uniform\PYZus{}initializer}\PY{p}{(}\PY{n}{minval}\PY{o}{=}\PY{o}{\PYZhy{}}\PY{l+m+mf}{0.003}\PY{p}{,} \PY{n}{maxval}\PY{o}{=}\PY{l+m+mf}{0.003}\PY{p}{)}
        \PY{n}{state\PYZus{}input} \PY{o}{=} \PY{n}{layers}\PY{o}{.}\PY{n}{Input}\PY{p}{(}\PY{n}{shape}\PY{o}{=}\PY{n+nb+bp}{self}\PY{o}{.}\PY{n}{state\PYZus{}size}\PY{p}{)}
        \PY{n}{l} \PY{o}{=} \PY{n}{layers}\PY{o}{.}\PY{n}{Dense}\PY{p}{(}\PY{l+m+mi}{128}\PY{p}{,} \PY{n}{activation}\PY{o}{=}\PY{l+s+s1}{\PYZsq{}}\PY{l+s+s1}{relu}\PY{l+s+s1}{\PYZsq{}}\PY{p}{)}\PY{p}{(}\PY{n}{state\PYZus{}input}\PY{p}{)}
        \PY{n}{l} \PY{o}{=} \PY{n}{layers}\PY{o}{.}\PY{n}{Dense}\PY{p}{(}\PY{l+m+mi}{64}\PY{p}{,} \PY{n}{activation}\PY{o}{=}\PY{l+s+s1}{\PYZsq{}}\PY{l+s+s1}{relu}\PY{l+s+s1}{\PYZsq{}}\PY{p}{)}\PY{p}{(}\PY{n}{l}\PY{p}{)}
        \PY{n}{l} \PY{o}{=} \PY{n}{layers}\PY{o}{.}\PY{n}{Dense}\PY{p}{(}\PY{l+m+mi}{64}\PY{p}{,} \PY{n}{activation}\PY{o}{=}\PY{l+s+s1}{\PYZsq{}}\PY{l+s+s1}{relu}\PY{l+s+s1}{\PYZsq{}}\PY{p}{)}\PY{p}{(}\PY{n}{l}\PY{p}{)}
        \PY{n}{net\PYZus{}out} \PY{o}{=} \PY{n}{layers}\PY{o}{.}\PY{n}{Dense}\PY{p}{(}\PY{n+nb+bp}{self}\PY{o}{.}\PY{n}{action\PYZus{}size}\PY{p}{[}\PY{l+m+mi}{0}\PY{p}{]}\PY{p}{,} \PY{n}{activation}\PY{o}{=}\PY{l+s+s1}{\PYZsq{}}\PY{l+s+s1}{tanh}\PY{l+s+s1}{\PYZsq{}}\PY{p}{,}
                               \PY{n}{kernel\PYZus{}initializer}\PY{o}{=}\PY{n}{last\PYZus{}init}\PY{p}{)}\PY{p}{(}\PY{n}{l}\PY{p}{)}
        \PY{n}{net\PYZus{}out} \PY{o}{=} \PY{n}{net\PYZus{}out} \PY{o}{*} \PY{n+nb+bp}{self}\PY{o}{.}\PY{n}{upper\PYZus{}bound}
        \PY{n}{model} \PY{o}{=} \PY{n}{keras}\PY{o}{.}\PY{n}{Model}\PY{p}{(}\PY{n}{state\PYZus{}input}\PY{p}{,} \PY{n}{net\PYZus{}out}\PY{p}{)}
        \PY{n}{model}\PY{o}{.}\PY{n}{summary}\PY{p}{(}\PY{p}{)}
        \PY{k}{return} \PY{n}{model}
\end{Verbatim}